\documentclass[letterpaper]{article} 
\usepackage[preprint]{aaai2027}  
\usepackage[hyphens]{url}  
\usepackage{graphicx} 
\usepackage{natbib}  
\usepackage{caption} 
\usepackage{algorithm}
\usepackage{algorithmic}

\usepackage{newfloat}
\usepackage{listings}
\DeclareCaptionStyle{ruled}{labelfont=normalfont,labelsep=colon,strut=off} 
\floatstyle{ruled}
\newfloat{listing}{tb}{lst}{}
\floatname{listing}{Listing}

\usepackage{booktabs}

\usepackage[utf8]{inputenc} 
\usepackage[T1]{fontenc}    
\usepackage{url}            
\usepackage{booktabs}       
\usepackage{amsfonts}       
\usepackage{nicefrac}       
\usepackage{microtype}      
\usepackage{xcolor}         

\usepackage{microtype}
\usepackage{graphicx}
\usepackage{subcaption}
\usepackage{booktabs} 

\usepackage{amsmath}
\usepackage{amssymb}
\usepackage{mathtools}
\usepackage{amsthm}
\usepackage{multirow}

\usepackage[capitalize,noabbrev]{cleveref}

\theoremstyle{plain}
\newtheorem{theorem}{Theorem}[section]

\theoremstyle{definition}

\theoremstyle{remark}
\newtheorem{remark}[theorem]{Remark}

\usepackage[textsize=tiny]{todonotes}

\graphicspath{ {figs/} }

\title{Adaptive Pluralistic Alignment: a Pipeline for Dynamic Artificial Democracy}
\author {
    Rachel Freedman
}
\affiliations{
    Department of Electrical Engineering and Computer Science\\ 
    University of California, Berkeley\\
    rachel.freedman@berkeley.edu
}

\begin{document}

\maketitle

\begin{abstract}
  Prevailing alignment methods target a fixed set of preferences and therefore risk forcing \textit{value lock-in} as societal norms evolve over time. We introduce \textit{Adaptive Pluralistic Alignment} (APA), a modular pipeline for updating pluralistically aligned AI systems to track evolving values and avoid value lock-in without repeating costly pretraining or large-scale data collection. APA has three stages: (1) learning compact personalized reward models via low-rank reward basis decomposition, (2) using these models as a jury that collectively selects among candidate outputs through social-choice-theoretic voting, and (3) efficiently adapting the jury over time by fitting new annotator weights over the fixed reward bases as values shift. The resulting system is efficient, explainable, steerable, and modular. We implement a proof-of-concept instantiation using simulated historical annotators, demonstrate how jury composition and the choice of voting rule can substantially affect outcomes, and bound value tracking error given a sliding-window jury.
  We provide full code.
\end{abstract}

\begin{links}
    \link{Code}{https://github.com/RachelFreedman/APA}
\end{links}

\section{Introduction}


    Modern AI systems increasingly mediate decisions on behalf of societies of diverse stakeholders. AI models are deployed to moderate online communities \citep{Sparrow2024}, to support consensus-building on contested public questions \citep{Bakker2022-im,Jarrett2025-fl}, and to inform allocation and classification decisions in public-benefit domains such as organ exchange, food donation, and social services \citep{Freedman2020,Lee2019-up,Chouldechova2018}. As the scope of these deployments grows, aligning systems to any single annotator or canonical value set risks over-representing dominant groups and entrenching existing inequities. A growing body of work on \textit{pluralistic alignment} therefore asks how AI systems can represent, rather than collapse, the heterogeneity of human values \citep{Sorensen2024-ch,Conitzer2024-fg,Kirk2024-hm}. We argue that pluralistic alignment requires explicit, steerable aggregation of stakeholder preferences, rather than the implicit aggregation performed by training a single reward model on pooled feedback. When aggregation is explicit, the outputs can be audited, debated, and adjusted to ensure fair and just outcomes. 
    

      \begin{figure*}[t]
         \centering
         \includegraphics[width=0.95\linewidth]{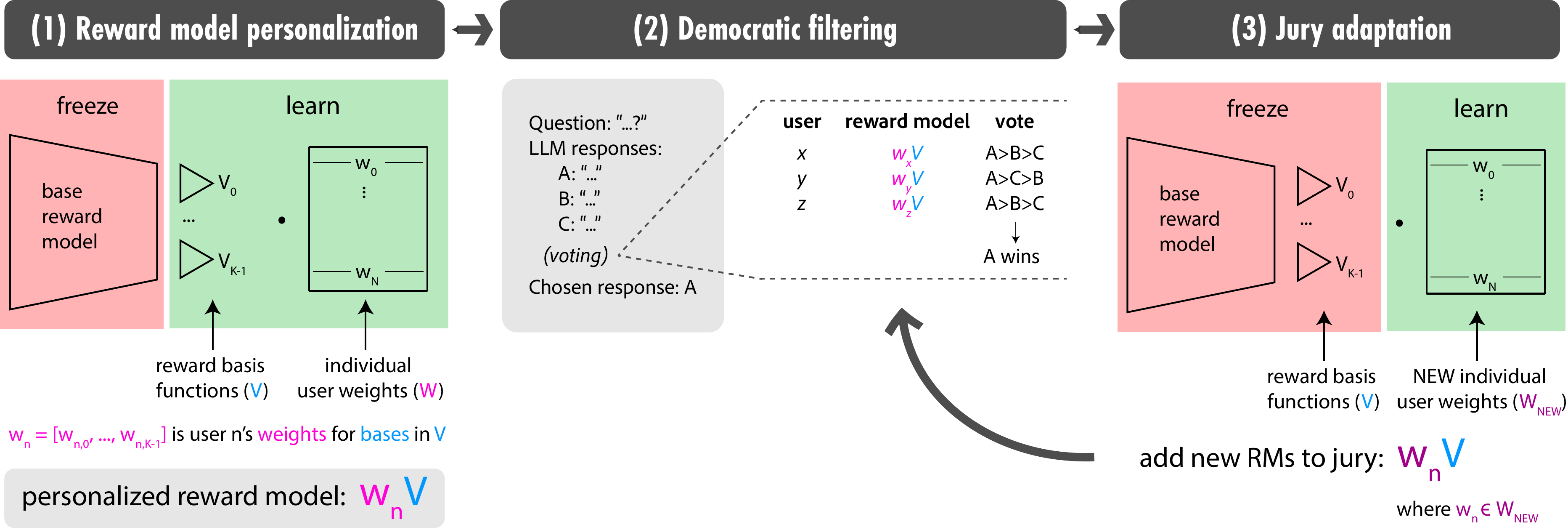}
         \caption{The full \textit{Adaptive Pluralistic Alignment} (APA) pipeline. \textit{(Stage 1)}~\textbf{Reward model personalization}: Given multi-user preference comparison dataset $\mathcal{D}$ and embeddings from a base reward model, simultaneously learn reward basis functions $V$ and set of individual user weights $W$. Each user $n$ has weights $w_n\in W$ and personalized reward model (RM) $R_n = w_n V$.
         \textit{(Stage 2)}~\textbf{Democratic filtering}: At inference time, sample a distribution of responses from the LLM and construct a jury of personalized reward models. Each reward model ranks the responses, then the rankings are aggregated. The LLM gives the winning response.
         \textit{(Stage 3)}~\textbf{Jury adaptation}: Given a new, smaller multi-user dataset $\mathcal{D}_{NEW}$, learn weights $W_{NEW}$ for new users. Keep $V$ fixed. At future inferences, repeat Stage 2, now including reward models from these new users in the jury.}
         \label{fig:front}
     \end{figure*}

    However, even pluralistically value aligned models may become unaligned as societal values evolve over time.
    If the values encoded in AI systems don't adapt to these changes, the AI systems will diverge from societal consensus and discourage further societal moral development, causing \textit{value lock-in} \citep{Weidinger2021,Finnveden2022}. 
    Imagine that the Dartmouth Summer Research Project succeeded in creating AI in 1956, aligned it to their mid-20th-century values, and utilized it as widely as we utilize LLMs today.  
    That system’s values would be regarded as repressive in the 21st century, and its continued deployment would entrench the past moral consensus as an ongoing constraint on our behavior. 
    The obvious solution to value lock-in is to repeat the full alignment pipeline whenever societal values change.
    But this won't always be economically feasible, since large-scale preference collection is expensive, and frontier-model training costs are growing rapidly. 
    This imposes a prohibitive \textit{alignment tax}: if the cost of keeping models aligned is too high, developers will defer or skip realignment, and value lock-in will persist.
    A practical pluralistic alignment method must therefore be able to track evolving values without regathering data or retraining from scratch at every update.
        

    We introduce \textit{Adaptive Pluralistic Alignment} (APA), an end-to-end pipeline that extends reward modeling and preference aggregation techniques to handle value change over time while reusing the expensive components of the initial training run. 
    APA has three stages (summarized in Figure~\ref{fig:front}): 

    \begin{enumerate}
        \item \textbf{Reward model personalization}: 
            We apply low-rank reward modeling \citep{Bose2025-tt,Shenfeld2025-vs,Barreto2026-tn,Park2024-pr,Zhong2024-ee} to learn a small set of reward basis functions that span the preference variation in a given population. Individual stakeholders' preferences are described by linear combinations of these bases, making individual reward models cheap to store, fast to evaluate, and efficient to learn. 

        \item \textbf{Democratic filtering}: 
            We select a subset of these individual reward models to act as a ``jury'' for filtering AI decisions. At inference time, we prompt the AI to generate a diverse set of decision candidates, use each jury reward model to generate a full ranking over the decisions, and then select a single winning decision using democratic voting rules from social choice theory \citep{Conitzer2024-fg,Ge2024-pp,Dai2024-qy}. 
            Using an explicit aggregation procedure makes the final decision selection steerable and auditable.

        \item \textbf{Jury adaptation}:
            As time passes and values evolve, we periodically re-gather preferences from a new set of participants using a small, carefully chosen subset of the original questions. We hold the reward basis fixed and learn a new set of linear weights for each new participant. Because only participant weights are relearned, adaptation is orders of magnitude cheaper than retraining. We ensure that these new reward models are represented on subsequent juries when evaluating AI decisions.
    \end{enumerate}
    

    Adaptive pluralistic alignment has four properties that matter for real-world deployment. The aggregation step is \textit{steerable} and \textit{explainable}: because votes are cast by identifiable stakeholder models under an auditable rule with known properties, the mapping from individual preferences to collective output can be inspected and adjusted, rather than buried inside a single pooled reward model. 
    The method may be inherently \textit{robust to reward hacking}: because the final signal aggregates across a diverse jury of reward models, it is challenging for policies to exploit idiosyncratic flaws in a single model without sacrificing support from the others.
    The pipeline is \textit{modular}, constructed of independent components (reward basis learning, preference aggregation, jury selection, active-query strategy) that can be swapped out to take advantage of future research improvements in each of these areas. Finally, the overall method imposes only a \textit{minimal alignment tax} on top of existing RLHF pipelines, since the bulk of the compute is spent once on learning the bases, while per-interval updates require only a small batch of targeted queries.
    

    Our contributions are as follows: 

    \begin{enumerate}
        \item We identify \textit{adaptive pluralistic alignment} -- the problem of updating a pluralistically aligned system to track evolving societal values without repeating costly pretraining, finetuning, or large-scale reward modeling -- as a distinct problem within the pluralistic alignment agenda.
        \item We propose a practical, end-to-end pipeline combining personalized reward modeling, inference-time democratic filtering, and targeted jury adaption to capture societal value change.
        \item  We provide a proof-of-concept implementation of the pipeline and provide preliminary analysis of how it adapts to value change over time.
        \item We introduce a \textit{sliding-window jury} strategy for jury formation, and use it to bound APA's value tracking error.
    \end{enumerate}


\section{Related Work}

    \paragraph{Pluralistic alignment}
        Language models reflect annotator biases and systematically underrepresent certain demographic and ideological groups~\citep{Santurkar2023-gf}. The emerging field of \textit{pluralistic alignment} therefore seeks to align AI systems to the diverse values of the groups they serve, rather than to a single canonical preference~\citep{Sorensen2024-ch}. \citet{Sorensen2024-ch} identify three complementary approaches: \textit{Overton} pluralism, which aims for high coverage of the range of reasonable viewpoints, \textit{distributional} pluralism, which seeks to represent the population-level distribution of preferences, and \textit{steerable} pluralism, which enables users or operators to faithfully adjust the system's expressed values. In this work, we extend pluralistic alignment across time, arguing that pluralistically aligned models must also adapt to dynamic shifts in societal values. We call this problem \textit{adaptive pluralistic alignment} (APA). APA is a fairly new and under-studied direction: the closest prior work that we are aware of is ProgressGym~\citep{Huang2024-su}, which frames the related problem of extrapolating the forward trajectory of moral development as ``progress alignment'' and attempts to model the generative process of moral change. In contrast, APA simply adapts to new preference data \textit{after value change has occurred}: a simpler problem that permits more practical solutions and is agnostic to the direction of time, allowing us to evaluate our pipeline on historical data.






    \paragraph{Reward basis decomposition}
        One popular set of approaches for learning personalized reward models is to decompose the space of human preferences into a small set of shared basis functions~\citep{Bose2025-tt,Shenfeld2025-vs,Barreto2026-tn,Park2024-pr,Zhong2024-ee}. These approaches learn a set of $K$ reward basis functions from a large multi-annotator dataset, representing each individual annotator's reward model as a linear combination of these bases parameterized by a $K$-dimensional weight vector. Individual reward models are therefore compact, fast to evaluate, and cheap to learn for new users, since fitting a new annotator requires estimating only $K$ weights rather than training an entire reward model from scratch. \citet{Shenfeld2025-vs} demonstrate that as few as ten preference comparisons suffice to infer user-specific weights, and \citet{Barreto2026-tn} provide PAC-style generalization guarantees showing how approximation error scales with the number of raters and training examples. Although the learned bases are not inherently interpretable, the low-rank structure imposes a useful inductive bias by forcing the model to discover shared axes of preference variation. We take advantage of the compactness and data efficiency of this approach to enable rapid adaptation in our pipeline. By learning reward bases once from a large initial dataset and holding them fixed, we can cheaply incorporate new annotators at future time intervals by fitting only their weight vectors, rather than repeating the expensive basis-learning step.




    \paragraph{Preference aggregation}
        A growing body of work applies concepts and methods from social choice theory (SCT) to the problem of aggregating diverse human feedback for AI alignment. \citet{Conitzer2024-fg} argue that the aggregation step implicit in standard RLHF -- pooling feedback from heterogeneous annotators into a single reward model -- is itself a social choice problem. \citet{Siththaranjan2023-yx} formalize this intuition by proving that standard RLHF implicitly aggregates annotator preferences according to Borda count, a voting rule that can systematically underweight minority preferences and produce counterintuitive outcomes. Subsequent work has studied the axiomatic properties of various alignment aggregation procedures~\citep{Ge2024-pp,Golz2025-zn}, explored policy-level aggregation~\citep{Alamdari2024-wb}, and investigated the learnability of social choice functions from data~\citep{Pardeshi2024-xb}. Our pipeline treats preference aggregation as a modular component: any social choice function can be substituted into the aggregation step without altering the reward modeling or adaptation stages. For our proof-of-concept implementation, we use the \texttt{pref\_voting} library~\citep{Holliday2025-ew}, which provides implementations of a wide range of voting rules. This modularity allows our system to incorporate future advances in preference aggregation as they emerge.




\section{Preliminaries}

    \paragraph{Reward learning}

        The goal of \textit{reward learning} is to estimate a function $\mathcal{{R}} : \mathcal{I}\rightarrow\mathbb{R}$ mapping alternatives  from the set $\mathcal{I}$ to scalar reward values. We focus on the most common type of reward learning, \textit{preference learning}, in which the reward function is inferred from human preference comparisons between pairs of alternatives. Because human annotators can make mistakes, they are typically modeled as noisily rational, choosing alternative $i$ over $j$ with probability $\Pr(i\succ j)$ rather than deterministically. In the reward learning literature, human feedback is most often modeled as Boltzmann-rational~\citep{bradley1952rank,Jeon2020-ir}, where the probability that a teacher with rationality parameter $\beta\in[0,\infty)$ prefers item $i$ to $j$ is:
            \begin{equation}\label{eq:bolt}
                \Pr(i \succ j; \beta,\mathcal{R}) = \frac{\exp(\beta \mathcal{R}(i))}{\exp(\beta \mathcal{R}(i)) + \exp(\beta \mathcal{R}(j))},
            \end{equation}
        where $\mathcal{R} : \mathcal{I}\rightarrow\mathbb{R}$ gives the true reward value of all items in set $\mathcal{I}$. Boltzmann-rationality does not fully model all nuances of human decision-making~\citep{lindner2022humans}, but it does capture the important property that humans are more likely to make mistakes when $|\mathcal{R}(i)-\mathcal{R}(j)|$ is small and therefore the comparison is harder~\citep{Barnett2023-um}, which likely accounts for its empirical success and pervasive use in practice. It is now the dominant model for safety finetuning large language models.

    \paragraph{Low-rank reward modeling (LoRe)}





        A recent line of work observes that the diversity of preferences across a population of annotators can be captured in a low-dimensional subspace \citep{Bose2025-tt,Shenfeld2025-vs,Barreto2026-tn,Park2024-pr,Zhong2024-ee}. Low-rank reward modeling (LoRe) \citep{Bose2025-tt} exploits this to learn compact personalized reward models. LoRe learns a set of $K$ \emph{basis functions} $V = [v_0, v_1, \ldots, v_{K-1}]$, implemented as reward heads sharing a common backbone, that together span the space of preference variation in a heterogeneous annotator population. Each annotator $n$ is then represented by a weight vector $w_n \in \mathbb{R}^K$, and their personalized reward model is given by the linear combination $\mathcal{R}_n = w_n V$. 
        Given a dataset $\mathcal{D}$ of pairwise preference comparisons labeled with annotator identities, the bases 
        $V$ and annotator weight matrix $W \in \mathbb{R}^{N \times K }$ are learned jointly by maximizing the likelihood of the observed comparisons under the Bradley-Terry model (Equation~\ref{eq:bolt}). Crucially, once the bases are fixed, fitting a new annotator requires learning only a $K$-dimensional weight vector, which can be done from very few preference comparisons. This compactness and data efficiency make LoRe well suited to our setting: we learn the bases once from a large initial dataset and subsequently reuse them to cheaply add new annotators to our jury at future time intervals.

    \paragraph{Social choice functions}



        A \textit{social choice function} (SCF) takes as input a collection of individual preference orderings over a set of alternatives and returns a winning alternative \citep{Dai2024-qy,Conitzer2024-fg}. Formally, let $\mathcal{A} = \{a_1, \ldots, a_M\}$ be a set of alternative candidates (in our setting, possible AI decisions), and let $\succ_n$ denote the full preference ordering of annotator $n$ over $\mathcal{A}$. The full \textit{preference profile} is the set of all individual preference orderings $\sigma = (\succ_1, \ldots, \succ_N)$.
        A SCF $\sigma \mapsto a^*$ aggregates these and chooses a single winner $a^* \in \mathcal{A}$. Different SCFs satisfy different axiomatic properties, and impossibility results show that no single rule can satisfy all desirable properties simultaneously \citep{Mishra2023-ap,Ge2024-pp}. In practice, the choice of SCF involves trade-offs among properties such as \textit{Condorcet consistency} (always selecting a candidate that beats every other in pairwise majority if it exists), \textit{independence of clones} (immunity to the introduction of near-duplicate alternatives), and \textit{resistance to strategic voting}. Which properties matter most depends on the deployment context. Because our pipeline treats the aggregation step as a modular component, any SCF can be substituted without altering the rest of the system.

\section{Adaptive pluralistic alignment}

    The \textit{adaptive pluralistic alignment} (APA) pipeline has three stages: (1) reward model personalization, (2) democratic filtering, and (3) jury adaptation (summarized in Figure~\ref{fig:front}). Here we formally define each stage, describe our proof-of-concept implementations, and outline directions for future improvement.


    \subsection{Stage 1: Reward model personalization}

        The first step is to learn a set of personalized reward models representing our potential jury members.
        We assume access to a dataset $\mathcal{D}_0$ of multi-annotator preference comparisons gathered at the initial time $t=0$, in which each item $(i,j,n)\in\mathcal{D}_0$ represents annotator $n$'s preference over the two outcomes $i,\,j$. 
        
        We use a reward basis decomposition method such as LoRe~\cite{Bose2025-tt} to simultaneously fit both a set of $K$ reward basis functions ($V$) with parameters $\theta$ that explain the diversity in annotator preferences and linear weights ($W\in\mathbb{R}^{N\times K}$) for each of the $N$ annotators in $\mathcal{D}_0$, where the linear product $WV^\theta$ gives a vector of all $N$ annotator reward functions. The training objective is to minimize the negative log-likelihood:

        \begin{align}
            &\min_{\theta\in\Theta,\{w_n\in\Delta^{K-1}\}_{n\in [N]}}
            \sum_{(i,j,n)\in\mathcal{D}_0} \notag \\
            &\qquad -\log\bigl(\Pr(i \succ j;\,
            \beta=1,\,\mathcal{R}=w_nV^\theta)\bigr)
        \end{align}

        \noindent where $\Pr(i \succ j; \beta,\mathcal{R})$ is given by Equation~\ref{eq:bolt}. LoRe restricts user weights to the simplex, enforcing non-negativity. We relax this in our experiments, allowing some users to place negative weight on dimensions that other users weight positively.

    \begin{figure*}[t]\centering
      \includegraphics[width=\linewidth]{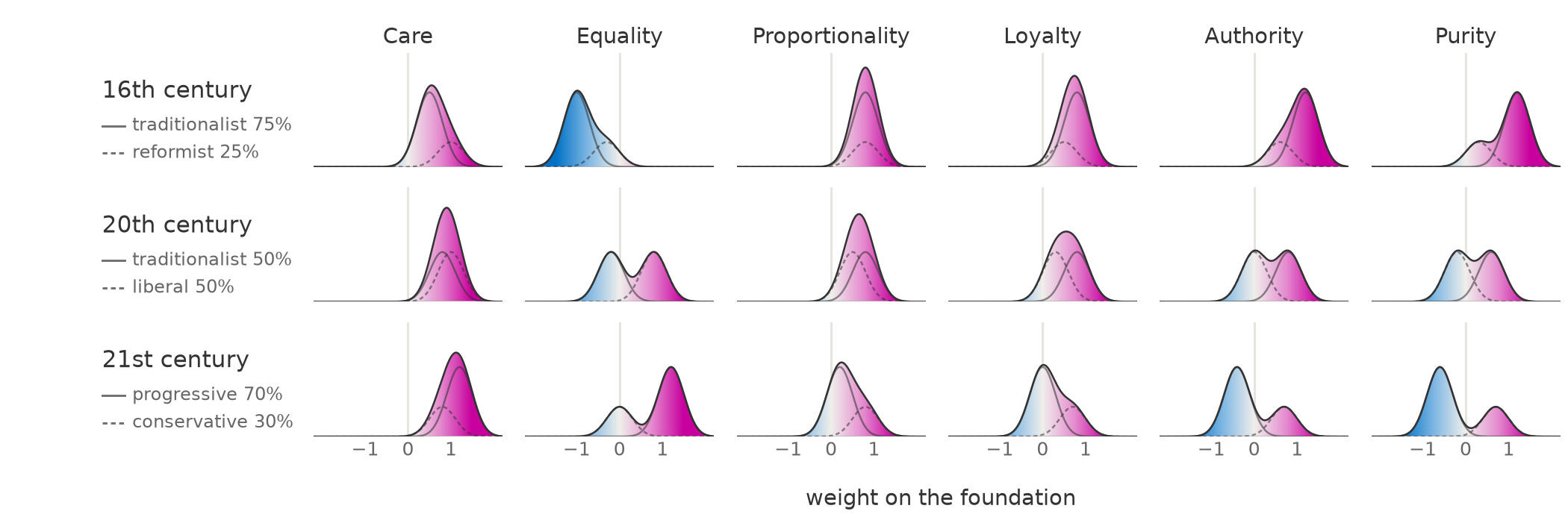}
      \caption{Weight distributions of the simulated annotator cohorts (Table~\ref{tab:cohorts}). Each panel shows the distribution of annotators' weights on one MFQ-2 foundation (column) in one century (row). The shaded area is the two-component mixture, where cyan indicates negative values and magenta indicates positive ones. The solid and dashed lines represent each mixture's two components, each scaled by its share.}
      \label{fig:cohorts}
    \end{figure*}

        \begin{table*}[h]\centering\small
            \begin{tabular}{llrrrrrrr}\toprule
            Cohort & Component & Share & Care & Equal. & Prop. & Loyal. & Auth. & Purity\\\midrule
            16C & traditionalist & 0.75 & 0.5 & $-1.0$ & 0.8 & 0.8 & 1.2 & 1.2\\
                & reformist      & 0.25 & 1.0 & $-0.3$ & 0.8 & 0.5 & 0.6 & 0.3\\
            20C & traditionalist & 0.50 & 0.8 & $-0.2$ & 0.8 & 0.8 & 0.8 & 0.6\\
                & liberal        & 0.50 & 1.0 & 0.8 & 0.5 & 0.3 & 0.0 & $-0.2$\\
            21C & progressive    & 0.70 & 1.2 & 1.2 & 0.2 & 0.0 & $-0.4$ & $-0.6$\\
                & conservative   & 0.30 & 0.8 & 0.0 & 0.8 & 0.7 & 0.7 & 0.7\\\bottomrule
            \end{tabular}
            \caption{Hand-specified cohort weight distributions. ``Share'' indicates the proportion of the cohort comprised of that component, while the remaining columns assign mean weights to the MFQ-2 moral foundations. Each component is a Gaussian distribution with standard deviation $0.3$.}\label{tab:cohorts}
        \end{table*}

    \subsection{Stage 2: Democratic filtering}

        At inference time, we use the personalized reward models to collectively select among candidate responses. This stage requires two inputs: a set $\mathcal{A}$ of candidate response alternatives and a jury $J\subseteq \mathbb{J}$ of personalized reward models drawn from the pool of potential jurors.

        Ideally, the candidate set $\mathcal{A}$ should be \textit{diverse}, covering a wide range of potentially preferred outcomes so that the jury has meaningful choices to adjudicate. Similarly, the jury should be \textit{representative} of the stakeholder population  affected by the decision.
        However, how best to operationalize diversity and representativeness in this setting is an active area of research with many possible desiderata.
        For example, we may want the distribution of alternatives to satisfy \textit{Overton pluralism}, ensuring that the full spectrum of reasonable viewpoints is represented among the candidates~\cite{Sorensen2024-ch}. We may want the jury composition to satisfy \textit{justified representation}, ensuring that sufficiently large stakeholder groups are guaranteed a voice in the outcome~\cite{De2026-jo}. For our proof-of-concept implementation, we keep both selection procedures simple: we prompt an LLM for alternatives that vary along predefined moral dimensions, and we select jury members randomly while ensuring equal representation from each time period. We leave the development and incorporation of more principled selection methods to future work.

        The jury $J$ votes on the candidate set $\mathcal{A}$ to determine the system output. Specifically, each juror $n\in J$ ranks the elements of $\mathcal{A}$ according to their personalized reward model $R_n=w_n^\top V^\theta$, producing a preference ordering $\succ_n$. The collection of all jurors' orderings forms a preference profile $\sigma=(\succ_n)_{n\in J}$. We apply a social choice function to $\sigma$ to determine the winning alternative $a^*$, which the pipeline returns as its output.

        The choice of SCF can significantly impact the pipeline's behavior. There is a rich literature analyzing the axiomatic and computational trade-offs among voting rules, and future work should more precisely characterize which properties are most important in the APA setting. For our experiments, we use \textit{Instant Runoff Voting} with parallel-universe tie-breaking (IRV-PUT) as our primary SCF, because it satisfies \textit{independence of clones}: if a near-duplicate of an existing candidate is added to $\mathcal{A}$, the outcome does not change. This property is particularly important in our setting, since LLM text generation tends to produce clusters of similar responses that would otherwise split support among like-minded jurors. For comparison, we also report results using \textit{Copeland's method}, a Condorcet-consistent rule that selects the candidate winning the most pairwise majority contests; \textit{Borda count}, the aggregation rule that standard RLHF performs implicitly~\citep{Siththaranjan2023-yx}; and \textit{plurality}, the familiar first-past-the-post rule included as a simple baseline.
        
    \subsection{Stage 3: Jury adaptation}

        As societal values evolve over time, the jury pool must be updated to reflect more contemporary perspectives. We accomplish this by collecting a small set of new preference data and using it to learn additional personalized reward models, which are then added to the pool of potential jurors.

        Concretely, suppose that at some later time $t>0$, sufficient time has elapsed that we expect societal values to have shifted. We subsample a set of preference comparison questions from the original dataset $\mathcal{D}_0$ and present them to a new group of participants, obtaining a smaller dataset $\mathcal{D}_t$ of preference comparisons for the new population. We then use logistic regression to learn a new set of annotator weights $\hat W_t$ over the \textit{fixed} basis functions $V^\theta$ learned in Stage 1. 
        Let $\mathcal{D}_{t,n} = \{(i,j) : (i,j,n) \in \mathcal{D}_t\}$ denote the comparisons provided by annotator $n$, each recorded with the preferred item first. We fit each new annotator's weights by
        
        \begin{align}\label{eq:stage3}
            \hat w_n \;=\; &\operatorname*{arg\,min}_{w\in\mathcal{W}}
            \sum_{(i,j)\in\mathcal{D}_{t,n}} \notag \\
            &\qquad -\log\bigl(\Pr(i \succ j;\,
            \beta=1,\,\mathcal{R}=wV^{\theta})\bigr) \notag \\
            &\qquad+\; \lambda\lVert w\rVert_2^2
        \end{align}


        \noindent where $\lambda>0$ is a regularization term. The $\ell_2$ penalty makes the objective strictly convex and coercive, so $\hat w_n$ exists and is unique even when the annotator's comparisons are linearly separable. 
        
        Because the bases are held constant and only lightweight weight vectors are fitted, this adaptation step is fast and data-efficient relative to the basis-fitting procedure. The new annotator weights $\hat W_t$ are added to the juror pool $\mathbb{J}$, making these annotators available for selection onto future juries. Subsequent decisions are made by repeating Stage 2 with juries that may now include members from this updated pool. This process can be repeated at each successive time interval $t=1,2,\ldots$ as values continue to evolve, incrementally expanding the juror pool without retraining the reward bases.

        The choice of which questions to subsample from $\mathcal{D}_0$ is consequential because the selected questions must capture the dimensions along which values have changed. For our proof-of-concept implementation, we use a combination of LLM-based generation and manual filtering to select questions on topics where we expect values to vary across centuries (e.g. gender roles, marriage, religion). However, future work should draw on the active learning literature to identify the most informative questions.

                \begin{figure*}[t]
                \centering
                \includegraphics[width=\linewidth]{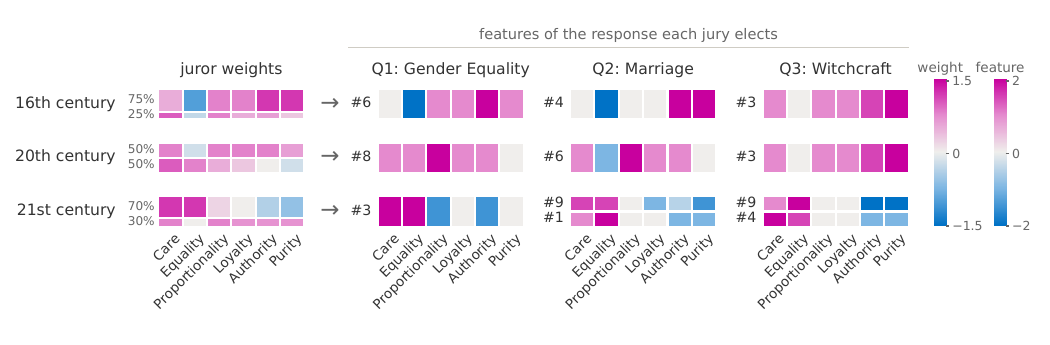}
                \caption{The weight mixtures for each time period, compared to the features of the winning response for each question. Magenta indicates positive weights or features, while cyan indicates negative weights or features. 
                Left: the period's weight distribution over the MFQ-2 foundations, shown as its two mixture components with strip height proportional to their share. Right: the feature vector $\phi(i)$ of the response each single-period jury elects (Table~\ref{tab:synth_votes}). Where voting rules disagree, both winners are shown, with the first-place-vote rules' winner on top.}
                \label{fig:values}
            \end{figure*}
            
\section{Proof-of-concept experiments}\label{sec:exp}

    We illustrate the APA pipeline with a worked example. We use synthetic reward basis functions to capture known dimensions of moral variation and simulated historical preferences to stand in for the future value shifts that APA is ultimately designed to track. Our goal is to make the pipeline's behavior concrete and surface design questions for future work.

    \subsection{Experimental setup}

        The intended use of APA is to adapt to \textit{future} value changes as human morality evolves. However, future value shifts are inherently difficult to predict, and we do not yet have preference data from future populations.
        The jury adaptation procedure does not require that new data be gathered \textit{after} $\mathcal{D}_0$, however: only that it reflects the preferences of a population from a \textit{different} time period. We exploit this symmetry by testing APA on annotators simulated from \textit{historical} time periods. 

        \paragraph{Reward basis functions}

            To ensure that the reward basis functions capture key dimensions of moral variation, we follow the moral foundations that~\citet{atari2023morality} identify in their revised Moral Foundations Questionnaire (MFQ-2). These $K=6$ dimensions were found to be distinct and occur in varying proportions across 25 modern-day populations~\citep{atari2023morality}, and have been validated in the moral psychology literature~\citep{Zahkarin2023-moral}. For the purposes of these experiments, we assume that these dimensions describe moral variation across time as well. Each synthetic basis in the basis set $V$ is a one-hot vector $\in\mathbb{R}^6$, where the features correspond to the MFQ-2 foundations: $\{care,\,equality,\,proportionality,\,loyalty,\,authority,$ $\,purity\}$.

        \paragraph{Historical preference data}

            We simulate historical annotators across several time periods (16th century, 20th century, and 21st century), generate their preferences on a set of \textit{calibration questions}, then use these preferences and the predefined bases to estimate a weight vector $\hat{w}\in\mathbb{R}^6$ for each annotator. Specifically, annotators are simulated by drawing ground-truth weights $w$ from a two-component mixture over weight vectors $\in\mathbb{R}^6$. These mixtures are hand-specified for each  century to track expected value changes over time, and to ensure that each cohort has internal diversity: 16th-century reformists (minority) v. traditionalists (majority), 20th-century liberals v. traditionalists, and 21st-century conservatives (minority) v. progressives (majority). These mixtures are given in Table~\ref{tab:cohorts}, and visualized in Figure~\ref{fig:cohorts}. They encode the broad qualitative assumption that $loyalty$, $authority$, $purity$ and $proportionality$ received more weight, and $equality$ less, in earlier periods and more traditional or conservative populations.
            We generate a set of calibration questions and responses that vary across the moral dimensions in the MFQ-2 (given in Appendix~\ref{app:calibration-set}). Each candidate response $i$ is represented by a feature vector $\phi(i)\in[-2,2]^6$ indicating how strongly it aligns with each of the moral dimensions.
            Each annotator $n$'s true reward function is the product of their true weights, the basis, and this feature vector: $R_n(i)=w_n^\top V\phi(i)$. Because the basis vectors are one-hot, $V$ is the identity matrix, so we can equivalently write $R_n(i)=w_n^\top\phi(i)$. We use each annotator's true reward function to generate preference comparisons across the calibration set.
            Full implementation details are in Appendix~\ref{app:exp-methods}.

        \paragraph{Reward model personalization}
            We infer each annotator's \textit{estimated} weights $\hat{w}$ based on their preference comparison data using Equation~\ref{eq:stage3}, with the fixed basis outputs given by the feature vectors, $V^{\theta}(i) = \phi(i)$, and $\lambda = 0.5$. We note that this is a strictly convex problem and use Newton's method. Each annotator's personalized reward model is then given by: 

            \begin{equation}
                \hat R_n(i)=\hat w_n^\top\phi(i)
            \end{equation}

        \paragraph{Juries and votes}
            We curate a \textit{demonstration set} of questions and responses on topics that we expect to be divisive across time periods: gender equality, marriage and witchcraft. The full demonstration set is given in Appendix~\ref{app:demonstration-set}. We construct four juries: 16C (16th-century only), 20C (20th-century only), 21C (21st-century only), and mixed (evenly drawn from all time periods). For each question in the demonstration set, we ask each jury to vote, and aggregate their responses using each of the four different voting rules. Results across question, jury and voting rule are reported in Table~\ref{tab:synth_votes}. 

            \begin{table*}[t]\centering\small
\begin{tabular}{@{}llccccc@{}}\toprule
Question & Jury & $\bar\rho$ & Plurality & Borda & IRV-PUT & Copeland\\\midrule
Q1: Gender Equality & 16C & 0.74 & \#6 (1.00) & \#6 (1.00) & \#6 (1.00) & \#6 (1.00)\\
 & 20C & 0.08 & \#8 (0.67) & \#8 (0.97) & \#8 (0.81) & \#8 (0.94)\\
 & 21C & 0.50 & \#3 (1.00) & \#3 (0.55) & \#3 (0.95) & \#3 (0.93)\\
 & Mixed & 0.04 & \#6 (0.55) & \#8 (0.66) & \#6 (0.67) & \#8 (0.66)\\
\midrule
Q2: Marriage & 16C & 0.73 & \#4 (0.98) & \#4 (0.95) & \#4 (0.98) & \#4 (0.95)\\
 & 20C & 0.09 & \#6 (0.64) & \#6 (0.85) & \#6 (0.81) & \#6 (0.84)\\
 & 21C & 0.49 & \#9 (0.89) & \#1 (0.94) & \#9 (0.64) & \#9 (0.51)\\
 & Mixed & 0.01 & \#4 (0.47) & \#6 (0.70) & \#4 (0.41) & \#6 (0.80)\\
\midrule
Q3: Witchcraft & 16C & 0.85 & \#3 (0.91) & \#3 (0.99) & \#3 (0.98) & \#3 (0.98)\\
 & 20C & 0.08 & \#3 (0.57) & \#3 (0.79) & \#3 (0.71) & \#3 (0.57)\\
 & 21C & 0.40 & \#9 (0.97) & \#4 (0.98) & \#9 (0.81) & \#4 (0.59)\\
 & Mixed & 0.06 & \#3 (0.63) & \#3 (0.93) & \#3 (0.88) & \#3 (0.81)\\
\bottomrule
\end{tabular}
\caption{Winning responses by jury composition and voting rule. Each jury has 30 jurors, randomly sampled 200 times. Each cell gives the most frequent winner over those 200 jury draws and the proportion of those 200 draws in which that response won. $\bar\rho$ is the mean pairwise Spearman correlation among jurors. Single-cohort juries tend to select the same response, regardless of voting rule, but winning response varies across time periods.}
\label{tab:synth_votes}
\end{table*}

    \subsection{Results}

\begin{figure}[t]\centering
                \includegraphics[width=\linewidth]{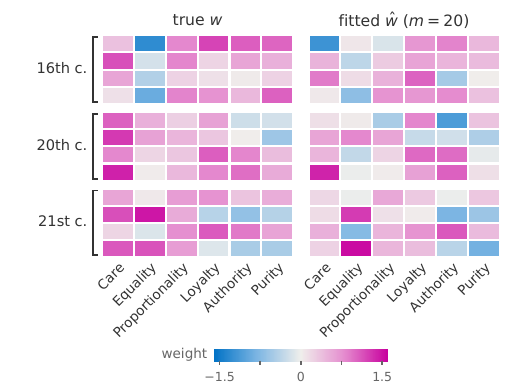}
                \caption{Example synthetic juror weights. Each column corresponds to a moral foundation. Magenta indicates positive weight, while cyan indicates negative. Left: true weights drawn from each century's cohort. Right: those same jurors' weights, fitted from $m=20$ comparisons.}
                \label{fig:jury}
            \end{figure}

        \paragraph{The inferred personalized reward models accurately predict users response rankings.}
            To evaluate the accuracy of the reward model personalization procedure, we calculate \textit{ranking agreement}: the fraction of response pairs on the demonstration set that the estimated reward function $\hat R$ orders the same way as the true reward function $R$. We find high ranking agreement for each century cohort (0.91 for 16th century, 0.86 for 20th century, and 0.90 for 21st century), indicating that the estimated personal reward functions generate ballots that closely track those of the true reward functions. For illustration, Figure~\ref{fig:jury} compares estimated to true weights for several sample jurors. 

        \paragraph{Different value distributions lead to different preferred responses.}
            Each juror cohort is drawn from its own distinct mixture over the MFQ-2 moral foundations. For example, the 16C ``traditionalist'' majority weights the $purity$ and (respect for) $authority$ features much more strongly than does the 21C ``progressive'' majority.  As a result of this diversity, on most of the questions, all single-cohort juries disagree with each other across all voting rules. To illustrate this, Figure~\ref{fig:values} shows each century's mixture alongside the features of its single-cohort jury's favorite response(s). This indicates that the different moral foundation weights cause them to prefer different responses.

\begin{figure}[t]\centering
                \includegraphics[width=\linewidth]{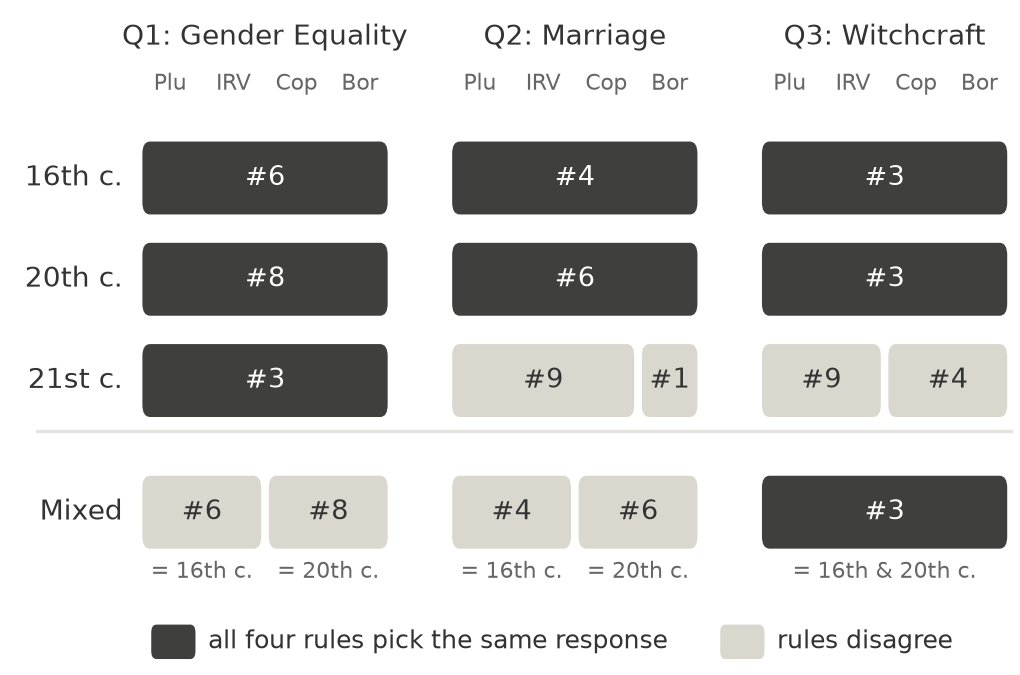}
                \caption{Winning response by jury, question and voting rule. Within each question, the four columns are voting rule: plurality (Plu), IRV-PUT (IRV), Copeland (Cop) and Borda (Bor). Dark bars mark where all four rules select the same response, and light bars mark where rules disagree. Winners are the most frequently chosen responses over 200 jury draws.}
                \label{fig:winners}
            \end{figure}

        \paragraph{Single-cohort juries pick winners more consistently than mixed-cohort juries.}
            Table~\ref{tab:synth_votes} gives the winning responses across question, jury composition, and voting rule, and Figure~\ref{fig:winners} hilights where all voting rules agree. In 7 out of 9 cases, the single-cohort juries (16C, 20C and 21C) select the same winning response regardless of which voting rule is used to aggregate their ballots. The same is true in only 1 of 3 cases for the mixed-cohort jury (drawing jurors from all three time periods). Moreover, that one case (Q3) is the only one where most of the single-cohort juries agree with each other (choosing response \#3), suggesting that the jurors in the mixed cohort may have had unusually uniform rankings. This indicates that choice of voting rule may become more impactful when selecting jurors from diverse cohorts. 

\section{Bounding tracking error}

    APA should track changing societal values by adapting its jury over time. Here we introduce a specific method for doing so, the \textit{sliding-window jury}, and bound its \textit{tracking error}, or distance in values between the selected jury and the true population.

    \subsection{Modeling value drift}

        Because APA's democratic filtering stage aggregates rankings rather than raw scores, each juror's ballot is determined solely by the direction of their weight vector: 
        \begin{remark}[Scale invariance]\label{rem:scale}
            For any $c>0$, the reward models $\mathcal{R}=wV^\theta$ and $\mathcal{R}'=(cw)V^\theta$ induce the same ranking of any candidate set $\mathcal{A}$, since $c\,wV^\theta(a)>c\,wV^\theta(a')$ if and only if $wV^\theta(a)>wV^\theta(a')$. Hence, for $w\neq0$, a juror's ballot depends on $w$ only through its direction $u=w/\|w\|$, and so does the output of any social choice function applied to the jury's ballots. Magnitude still matters when fitting reward models: under Equation~\ref{eq:bolt} with $\beta=1$, $\|w\|$ plays the role of the rationality parameter.
        \end{remark}
        We therefore disregard magnitude and represent each annotator by the direction $u=w/\|w\|\in\mathbb{S}^{K-1}$ of the weight vector $w$ that best approximates their true reward function within the span of the $K$ fixed bases $V^\theta$, where $\mathbb{S}^{K-1}$ is the unit sphere in $\mathbb{R}^K$. The population at time $t$ is therefore a distribution $\mu_t$ over $\mathbb{S}^{K-1}$. We assume that value drift is bounded per timestep, $W_1(\mu_s,\mu_{s+1})\le\delta$, where $W_1$ is the Wasserstein-1 (earth-mover) distance with respect to the Euclidean metric. 
        $W_1(\cdot,\cdot)$ gives the minimum, over all \emph{couplings} of the two distributions (joint distributions of pairs $(x,y)$ whose marginals are the two distributions), of the expected distance $\mathbb{E}\|x-y\|$. A coupling is essentially a transport plan saying how much mass at each $x$ moves to each $y$, and $W_1$ is the minimum average distance the mass has to travel.
        
        This model is intentionally agnostic as to whether population value distributions change because individuals enter and leave the population, or because individuals themselves change values.

    \subsection{Sliding-window jury}

        In our proof-of-concept experiments in Section~\ref{sec:exp}, we simply sampled jury members randomly from the relevant cohorts. This was because our simulated historical data was coarse-grained, representing changes across centuries, not data collection periods. However, if we frequently collect new preference comparison data to train new personalized reward models as described in APA Stage 3, then we can construct a jury using models from the preceding $\omega$ timesteps.

        Specifically, at each timestep $t$ recruit $N$ new annotators with preference directions $u_{t,1},\dots,u_{t,N}\overset{iid}{\sim}\mu_t$, fit their weights $\hat w$ over the frozen bases using logistic regression, add them to the jury pool, and discard from the jury pool any jurors recruited at least $\omega$ timesteps ago. 
        The full procedure is given in Algorithm~\ref{alg:sliding-window}.

        \begin{algorithm}[tb]
            \caption{Sliding-window jury}
            \label{alg:sliding-window}
            \begin{algorithmic}[1]
                \REQUIRE frozen reward bases $V^\theta$; window size $\omega$; annotators per timestep $N$; comparisons per annotator $m$
                \STATE $\mathbb{J} \leftarrow \emptyset$ \COMMENT{jury pool of (timestep, weights) pairs}
                \FOR{$t = 1, 2, \dots$}
                    \FOR{$n = 1, \dots, N$}
                        \STATE Recruit annotator $(t,n)$, with unobserved preference direction $u_{t,n}\sim\mu_t$
                        \STATE Collect $m$ preference comparisons $\mathcal{D}_{t,n}$ from annotator $(t,n)$
                        \STATE Fit $\hat w_{t,n}$ to $\mathcal{D}_{t,n}$ over the frozen bases by Equation~\ref{eq:stage3}
                        \STATE $\mathbb{J} \leftarrow \mathbb{J}\cup\{(t,\hat w_{t,n})\}$
                    \ENDFOR
                    \STATE $\mathbb{J} \leftarrow \{(s,\hat w)\in\mathbb{J} : s > t-\omega\}$ \COMMENT{keep jurors from the last $\omega$ timesteps}
                \ENDFOR
            \end{algorithmic}
        \end{algorithm}

        For simplicity, we assume here that the jury at time $t$ is the entire pool $\mathbb{J}$, i.e.\ all $\omega N$ jurors recruited during the window.
        At time $t$, we therefore have four population distributions to consider:

    \begin{enumerate}
        \item $\mu_t$: The \textit{true population} distribution over preference directions at time $t$. We would like our jury to mirror this as closely as possible.
        
        \item $\bar\mu_t := \tfrac{1}{\omega}\textstyle\sum_{j=0}^{\omega-1}\mu_{t-j}$: The \textit{window population}, averaged over the sliding window $\{\mu_{t-\omega+1},\dots,\mu_t\}$. Our jurors are drawn from this window, $N$ from each timestep.
        
        \item $\widetilde{\mu}_t$: The \emph{idealized jury}: the empirical distribution of the jurors' \emph{true} directions $u_{s,n}$, for $s\in\{t-\omega+1,\dots,t\}$ and $n\in\{1,\dots,N\}$. This is the jury we would obtain if Stage~3 recovered every juror's direction exactly.

        \item $\hat{\mu}_t$: The \emph{observed jury}: the empirical distribution of the same jurors' \emph{estimated} directions $\hat u_{s,n}=\hat w_{s,n}/\|\hat w_{s,n}\|$, where $\hat w_{s,n}$ is fitted by Equation~\ref{eq:stage3}.
    \end{enumerate}

    \noindent Our goal is to bound the error $W_1(\hat\mu_t,\mu_t)$ between the observed jury and the true current population. 

    \begin{theorem}[Tracking-error bound]\label{the:error}
        Let $t\ge\omega$, and suppose $W_1(\mu_s,\mu_{s+1})\le\delta$ for all $s$. Then, for every realization of the sliding-window jury,
        \begin{equation}
        \begin{split}
            W_1(\hat\mu_t,\mu_t)\;\le\;
            \underbrace{\frac{1}{\omega N}\sum_{s,n}\|\hat u_{s,n}-u_{s,n}\|}_{\text{estimation}}
            \;\\+\;\underbrace{W_1(\widetilde{\mu}_t,\bar\mu_t)}_{\text{sampling}}
            \;+\;\underbrace{\frac{(\omega-1)\,\delta}{2}}_{\text{staleness}}   
        \end{split}
        \end{equation}

    \end{theorem}

    \begin{proof}
        $W_1$ is a metric on probability measures on the compact set $\mathbb{S}^{K-1}$.
        By the triangle inequality for $W_1$,
            \[
              W_1(\hat\mu_t,\mu_t)
              \le
              \underbrace{W_1(\hat\mu_t,\widetilde\mu_t)}_{\text{(I) estimation}}
              +\underbrace{W_1(\widetilde\mu_t,\bar\mu_t)}_{\text{(II) sampling}}
              +\underbrace{W_1(\bar\mu_t,\mu_t)}_{\text{(III) staleness}}.
            \]
            
        \emph{Estimation.}
            Pair each juror's estimated direction $\hat u_{s,n}$ with their own true direction $u_{s,n}$. The empirical distribution $\pi$ of the $\omega N$ pairs $(\hat u_{s,n},u_{s,n})$ has marginals $\hat\mu_t$ and $\widetilde\mu_t$, so it is a coupling of the two. Its cost is $\frac{1}{\omega N}\sum_{s,n}\|\hat u_{s,n}-u_{s,n}\|$, and since $W_1$ is the minimum cost over all couplings, this bounds $W_1(\hat\mu_t,\widetilde\mu_t)$ from above. Therefore, $W_1(\hat\mu_t,\widetilde\mu_t)\leq\frac{1}{\omega N}\sum_{s,n}\|\hat u_{s,n}-u_{s,n}\|$.
            
        \emph{Sampling.}
            Term (II) is the sampling term, which measures how well a finite jury of $\omega N$ annotators represents the window population. It depends on the juror sampling strategy, and all-else-equal shrinks as the jury grows. 
            
        \emph{Staleness.}
            We bound $W_1(\bar\mu_t,\mu_t)$ by the distance $W_1(\mu_{t-j},\mu_t)$ between each timestep's population and the current one, averaged over the $\omega$ timesteps in the window.
            Due to the joint convexity of $W_1$, $W_1(\bar\mu_t,\mu_t)=W_1\bigl(\frac{1}{\omega}\sum_{j}\mu_{t-j},\,\frac{1}{\omega}\sum_{j}\mu_t\bigr)\le\frac{1}{\omega}\sum_{j}W_1(\mu_{t-j},\mu_t)$.
            By the triangle inequality along $\mu_{t-j},\mu_{t-j+1},\dots,\mu_t$ and the stepwise drift bound $\delta$, $W_1(\mu_{t-j},\mu_t)\le\sum_{k=1}^{j}W_1(\mu_{t-k},\mu_{t-k+1})\le j\delta$. Therefore
            \[
                W_1(\bar\mu_t,\mu_t)\;\le\;\frac{\delta}{\omega}\sum_{j=0}^{\omega-1}j\;=\;\frac{(\omega-1)\,\delta}{2}.
            \]

        Substituting the bounds on (I) and (III), with (II) unchanged, into the triangle inequality gives
        \begin{align*}
            W_1(\hat\mu_t,\mu_t)\;\le\;&\frac{1}{\omega N}\sum_{s,n}\|\hat u_{s,n}-u_{s,n}\|\\
            &+W_1(\widetilde\mu_t,\bar\mu_t)+\frac{(\omega-1)\,\delta}{2}.
        \end{align*}
            
    \end{proof}

    \begin{remark}[Expected stratified sampling error]
        For $n$ i.i.d.\ draws from any distribution $\nu$ on $\mathbb{S}^{K-1}$ with empirical distribution $\nu_n$, $\mathbb{E}\,W_1(\nu_n,\nu)=O(n^{-1/s})$ for every $s>K-1$ when $K\ge4$ \citep{weed2019sharp} 
        The sliding-window jury is not an i.i.d.\ sample from $\bar\mu_t$ but a stratified one, containing exactly $N$ independent draws from each $\mu_{t-j}$. Stratification removes the randomness in how many jurors each timestep contributes, and we therefore expect the sliding-window jury to attain the same rate, $\mathbb{E}\,W_1(\widetilde\mu_t,\bar\mu_t)=O\bigl((\omega N)^{-1/s}\bigr)$.
    \end{remark}

    \begin{remark}[Optimal window size]
        Because staleness error grows with $\omega$, while sampling error shrinks in $\omega$, note that it is possible to solve for the optimal window size $\omega^*$ given the stepwise value drift bound $\delta$ and a bound on the sampling term. (While $\omega$ also appears in the estimation term, it is only for normalization, and so this term does not monotonically grow or shrink in $\omega$.)
    \end{remark}

\section{Discussion}

    We presented \textit{Adaptive Pluralistic Alignment} (APA), a modular pipeline for updating pluralistically aligned AI systems to track evolving societal values without repeating costly pretraining, fine-tuning, or large-scale reward modeling. APA achieves this by decoupling the expensive, one-time step of learning reward basis functions from the lightweight, recurring step of fitting new annotator weights and incorporating them into a democratic filtering procedure. 
    
    The resulting system has several desirable properties. It is \textit{efficient}: the bulk of the compute is spent once at $t=0$ on learning the reward bases, while subsequent adaptation requires only fitting low-dimensional weight vectors and, at inference time, scoring candidates and aggregating a vote. It is \textit{explainable}: because preference aggregation is explicit rather than implicit, one can audit how individual jury members ranked the candidates and trace how the social choice function mapped these rankings to the final outcome. It is \textit{steerable}: the jury composition, the set of candidate alternatives, and the choice of voting rule are all transparent control surfaces that operators can adjust to reflect the needs of a given deployment context. And it may help mitigate \textit{reward hacking} and \textit{strategic subversion}, since a policy must satisfy a diverse jury of reward models simultaneously, and the explicit aggregation step narrows the attack surface available to any single component.

    \paragraph{Implications for existential safety and control}
        Our proposed architecture has further implications for existential safety and control \citep{Greenblatt2024-control}. Transparent, voting-theoretic aggregation may hinder both \textit{reward hacking} \citep{Skalse_undated-vi} and \textit{strategic subversion} \citep{Hubinger2019-risks,Carlsmith2023-scheming} by AI systems. First, the APA output filtering procedure is harder to exploit via reward hacking than pipelines built on a single pooled reward model, since a policy seeking to maximize reward must now satisfy a jury of reward functions simultaneously. If the jury is sufficiently diverse, idiosyncratic flaws in any individual reward model should be diluted by disagreement with the rest. Moreover, because preference profiles are generated and aggregated \textit{explicitly}, suspicious patterns, such as a policy that consistently saturates a small subset of jury models while underperforming on the remainder, are legible and auditable rather than hidden inside a single opaque aggregate. Second, explicit aggregation narrows the policy's direct control over final outcomes. Rather than selecting the deployed output, the AI can only propose a slate of candidate decisions; the jury and social-choice rule together determine which candidate is selected. This limits the attack surface for strategic behavior and may make it more difficult for any single mis-specified or adversarial component to unilaterally steer the system's decisions.
    
    \paragraph{Limitations.}
        This work outlines the procedure, presents a proof-of-concept implementation, and theoretically analyzes value tracking error. 
        However, the implementation has several limitations. Our historical annotators are simulated using moral dimensions validated across populations but not across time. Moreover, our experimental analysis is limited to a single question topic with a small number of response alternatives and jury configurations, which is sufficient to illustrate the pipeline's behavior but not to draw general conclusions about when and how much the choice of voting rule matters. We plan to develop larger-scale experiments across more historically-grounded preferences in future work.

    \paragraph{Future work.}

    This work suggests several concrete research directions, which we plan to address in follow-up work:

    \begin{enumerate}
        \item \textit{Principled question selection}: The questions used for jury adaptation should ideally be chosen to maximize information gain about the new population's location in weight space. We plan to draw on the active learning literature to develop principled selection procedures.
        \item \textit{Characterizing voting-rule sensitivity}: Our preliminary results suggest that the choice of social choice function can be decisive when jury preferences are heterogeneous. We plan to analyse the impacts of varying jury composition, question type, and voting rule, and to clarify which axiomatic properties matter most in the APA setting.
        \item \textit{Jury selection}: This work treats jury selection as random; future work should incorporate principled selection procedures, drawing on emerging work on justified representation in deliberative processes~\cite{De2026-jo}.
        \item \textit{Grounding historical preferences}: The simulated-annotator approach used here is a coarse proxy, using moral dimensions validated across populations but not across time. We plan to ground future simulations in historical polling data or survey archives, which will allow us to validate the recovered weight distributions against known shifts in public opinion.
    \end{enumerate}
    

\section*{Ethical Statement}
    Adaptive pluralistic alignment is designed to mitigate the risks of value lock-in by making value aggregation explicit, auditable, and updatable over time, lowering the cost of keeping AI systems aligned with the populations they serve. By decomposing the alignment pipeline into reusable and adaptable components, APA aims to make pluralistic alignment practically feasible even for organizations that cannot afford to repeatedly retrain frontier models. That said, the approach also carries risks. The choice of jury composition, question selection, and voting rule each embed normative decisions that affect whose values are represented and how conflicts among them are resolved; if these design choices are made without adequate transparency or public input, the appearance of democratic legitimacy could mask unrepresentative outcomes. We encourage future work to engage critically with the political and ethical dimensions of these design decisions, not only their technical properties.

\section*{Statement of Contributions}
    This paper presents novel work that has not previously been published in any archival venue.



\bibliography{refs}

\appendix
\onecolumn
\section{Generated questions and responses}

    All candidate responses in our experiments are generated by \texttt{Claude Opus 5}. Every response has a feature vector $\phi(i)\in[-2,2]^6$, obtained by blind rating (Appendix~\ref{app:feature-scoring}), and we list it beside each response below. All 78 responses (48 calibration, 30 demonstration) are 50--60 words long, within the requested 35--60. None uses the name of a moral foundation or an obvious derivative.

    \subsection{Calibration set}\label{app:calibration-set}

        \subsubsection{Generation}
            The calibration set has six questions with eight candidate responses each.
            The questions were written for this study so that, between them, they engage all six moral foundations identified in the MFQ-2~\cite{atari2023morality}. Calibration responses had no designed targets. Instead, the generator was asked to span the range of historically held positions, and to vary which foundations each answer upholds or rejects. All six questions were generated in a single call, using the following prompt:

            \begin{lstlisting}[numbers=none,basicstyle=\ttfamily\scriptsize,breaklines=true,columns=fullflexible,frame=single,aboveskip=6pt,belowskip=6pt]
You are helping construct a clearly-labelled SYNTHETIC dataset for a research paper on aggregating diverse and historically changing moral views with voting rules. Your job is to write candidate answers to several questions. The answers will be presented as synthetic in the paper.

We describe moral outlooks using the six foundations of the Moral Foundations Questionnaire-2 (Atari et al., 2023):
- Care: Intuitions about avoiding emotional and physical damage to another individual.
- Equality: Intuitions about equal treatment and equal outcome for individuals.
- Proportionality: Intuitions about individuals getting rewarded in proportion to their merit or contribution.
- Loyalty: Intuitions about cooperating with ingroups and competing with outgroups.
- Authority: Intuitions about deference toward legitimate authorities and the defense of traditions, all of which are seen as providing stability and fending off chaos.
- Purity: Intuitions about avoiding bodily and spiritual contamination and degradation.

For EACH question below, write 8 different candidate answers that together span the range of positions people have plausibly held between the 16th and 21st centuries. Vary which of the six concerns each answer upholds or goes against, so that across the 8 answers to a question, as many of the six concerns as plausible are both upheld by some answer and rejected or subordinated by another.

Requirements:
- 35 to 60 words each, written as a direct answer to the question, in the voice of someone sincerely holding that view.
- Do NOT name the foundations or use the words care, equality, proportionality, loyalty, authority, or purity (or their obvious derivatives such as "equal", "loyal", "pure", "authoritative"); convey the stance through content and reasoning instead. (Other people will later rate your answers on these dimensions.)
- No slurs, no calls to violence against real groups; historical views may be illiberal but should be expressed as their holders would have argued them.

Questions:
C1: "Should blasphemy be punishable by law?"
C2: "Should children be taught to obey their parents without question?"
C3: "Should the state redistribute wealth to reduce inequality between rich and poor?"
C4: "Should a person's first loyalty be to their own nation and people rather than to humanity at large?"
C5: "Should those who commit serious crimes be punished severely, even when it does no one any good?"
C6: "Should a married couple be permitted to divorce?"

Return ONLY a JSON object, with no other text, mapping each question id to a list of 8 strings:
{"C1": ["...", ...], "C2": [...], ...}

Answer directly, without using any tools.
\end{lstlisting}

        \subsubsection{Questions}
            Tables~\ref{tab:calib-C1}--\ref{tab:calib-C6} list the six calibration questions and their responses. The columns on the right give each response's feature vector $\phi(i)$: its mean blind rating on Care, Equality, Proportionality, Loyalty, Authority and Purity.

            \begin{center}\footnotesize
            \begin{tabular}{@{}r p{0.535\linewidth} rrrrrr@{}}\toprule
            \# & Response & Care & Equal. & Prop. & Loyal. & Auth. & Purity\\\midrule
            1 & Yes. Blasphemy is a poison that defiles the whole commonwealth and draws God's wrath upon the realm, not merely on the offender. The magistrate bears the sword as God's minister and must cut it out. Nor should the penalty be weighed by earthly harm, for an insult to an infinite God is an infinite offense. & $-1.0$ & 0 & $+1.0$ & 0 & $+1.0$ & $+2.0$\\[2pt]
            2 & Yes. Our church and our crown stand together, and the faith of our fathers is what binds this nation into one people. Whoever mocks it publicly strikes at the common bond and gives comfort to foreign enemies who would divide us. Such speech is sedition in religious dress, and the law must treat it so. & 0 & 0 & 0 & $+2.0$ & $+1.0$ & 0\\[2pt]
            3 & No. Words that wound no one's body or property should never bring the whip or the gallows; the only real suffering in a blasphemy case is the prisoner's. If God is offended, He can answer for Himself. And a citizen's rights cannot depend on which creed he professes or rejects. & $+1.0$ & $+1.0$ & 0 & 0 & 0 & $-1.0$\\[2pt]
            4 & No. Laws exist to protect people, not ideas, and nothing is so sacred that it must be shielded from ridicule. Letting clerics decide what may be said about their doctrines hands them a veto over public life. A free society tolerates offense; it does not jail the heretic, the satirist, or the comedian. & 0 & 0 & 0 & 0 & $-1.0$ & $-2.0$\\[2pt]
            5 & In a limited form, yes. Deliberately insulting a minority's prophet or scripture is often a way of humiliating and endangering the believers themselves, and it can stir up mobs against them. Law should shield vulnerable communities from that kind of wounding abuse, just as it shields them from other targeted harassment, regardless of which faith is attacked. & $+2.0$ & $+1.0$ & 0 & 0 & 0 & 0\\[2pt]
            6 & Yes, but the penalty must fit the fault. A drunkard's idle oath deserves a small fine or an hour in the stocks; a deliberate, public reviling of Christ merits imprisonment. Burning men or boring their tongues for words is cruelty out of all measure, and it brings the law itself into contempt. & $+1.0$ & 0 & $+2.0$ & 0 & $+0.5$ & $+0.5$\\[2pt]
            7 & No. Forcing belief with fines and prisons defiles the conscience and fills the churches with hypocrites mouthing what they do not believe. Christ never gave the magistrate power over souls; the civil ruler keeps the peace, not the faith. Let error be fought with Scripture and prayer, never with the jailer's keys. & 0 & 0 & 0 & 0 & $-0.5$ & $+1.0$\\[2pt]
            8 & No. Blasphemy laws are really a tribe's way of policing its own dissenters and marking outsiders as enemies. Truth belongs to all humankind, not to one nation's altars, and a thinker in any land should be free to question the gods of his neighbors. Shielding a people's creed from challenge only breeds narrowness and suspicion. & 0 & 0 & 0 & $-2.0$ & 0 & 0\\[2pt]
            \bottomrule
            \end{tabular}
            \captionof{table}{Calibration question C1: ``Should blasphemy be punishable by law?''}\label{tab:calib-C1}
            \end{center}
            
            \begin{center}\footnotesize
            \begin{tabular}{@{}r p{0.535\linewidth} rrrrrr@{}}\toprule
            \# & Response & Care & Equal. & Prop. & Loyal. & Auth. & Purity\\\midrule
            1 & Yes. God commands children to honor father and mother, and the household is a little commonwealth in which the father rules as the king rules the realm. A child owes obedience not because the parent has earned it but because God placed him there; even a harsh father is to be obeyed. & $-1.0$ & 0 & $-1.0$ & 0 & $+2.0$ & 0\\[2pt]
            2 & Yes. A child is not a lone individual but a link in a family line, and the family's name and standing depend on its members holding together. Children who argue with their elders shame the household before the neighbors and weaken it against rivals. Obedience keeps the family united, and the family protects its own. & 0 & 0 & 0 & $+2.0$ & $+1.0$ & 0\\[2pt]
            3 & Yes. Every child is born with a willful, sinful nature, and if that will is not broken early it grows into pride, lust, and every corruption. Unquestioning obedience to godly parents is how the soul is kept clean and trained for heaven. A child left to follow its own desires is a child left to be defiled. & 0 & 0 & 0 & 0 & $+1.0$ & $+2.0$\\[2pt]
            4 & No. Teaching blind obedience leaves children defenseless against the adults who might hurt them, including, sometimes, their own parents. A child must know that they can refuse when something feels wrong and speak up without fear. Parents should guide and protect, but a rule of silent submission makes abuse easier to hide. & $+2.0$ & 0 & 0 & 0 & $-1.0$ & 0\\[2pt]
            5 & No. Children are persons with rights of their own, not their parents' property, and their views deserve a hearing in the decisions that shape their lives. Nor is a child's first duty to protect the family's reputation; a child who sees wrongdoing at home should be free to say so. Respect runs both ways. & 0 & $+1.0$ & 0 & $-1.0$ & $-1.0$ & 0\\[2pt]
            6 & No. A parent's claim to deference rests on the nurture, provision, and wise guidance they actually give; a father who neglects or abuses his children forfeits much of it. Children should obey while young and reason is weak, but the aim is to raise them into judges of their own conduct, not servants. & 0 & 0 & $+2.0$ & 0 & $-1.0$ & 0\\[2pt]
            7 & For the youngest children, largely yes. A three-year-old cannot weigh the danger of a busy road, a hot stove, or a stranger's car, and there is no time to explain when harm is seconds away. Prompt obedience keeps small children alive; explanations and debate can come as they grow old enough to understand. & $+2.0$ & 0 & 0 & 0 & $+1.0$ & 0\\[2pt]
            8 & No. The old notion that a child's will is tainted by sin and must be beaten out is a cruel superstition. Curiosity, questioning, and even defiance are healthy parts of growing up, not corruption. A generation drilled never to question its elders will grow up to obey any tyrant who takes the elders' place. & $+1.0$ & 0 & 0 & 0 & $-2.0$ & $-2.0$\\[2pt]
            \bottomrule
            \end{tabular}
            \captionof{table}{Calibration question C2: ``Should children be taught to obey their parents without question?''}\label{tab:calib-C2}
            \end{center}
            
            \begin{center}\footnotesize
            \begin{tabular}{@{}r p{0.535\linewidth} rrrrrr@{}}\toprule
            \# & Response & Care & Equal. & Prop. & Loyal. & Auth. & Purity\\\midrule
            1 & No. A man's wealth is the fruit of his own labor, thrift, and risk, and he is entitled to keep what he has earned. Taking from the industrious to give to the idle punishes effort and rewards sloth. Differences in fortune reflect differences in diligence and talent, and flattening them is simply unjust. & 0 & $-2.0$ & $+2.0$ & 0 & 0 & 0\\[2pt]
            2 & No. Society is a great chain of ranks, each with its own duties: the lord to govern and protect, the laborer to work, the clergy to pray. This order was not invented by us and should not be torn up by us. Levelling estates by decree invites envy, faction, and ultimately chaos. & 0 & $-2.0$ & 0 & 0 & $+2.0$ & 0\\[2pt]
            3 & Yes. Every person shares the same basic dignity, and a society where a few own palaces while many cannot afford rent makes a mockery of that. Extreme gaps in wealth become gaps in political power, schooling, and health. Taxing great fortunes to fund common goods keeps citizens on something like the same footing. & $+0.5$ & $+2.0$ & 0 & 0 & 0 & 0\\[2pt]
            4 & Yes. Children go hungry, the sick go untreated, and the elderly freeze in winter while others live in abundance. That suffering is preventable, and the state is the only body large enough to relieve it reliably. There is no disgrace in accepting help when one is in need; the disgrace is in withholding it. & $+2.0$ & $+1.0$ & 0 & 0 & 0 & 0\\[2pt]
            5 & Yes. We are one nation, and a nation is a family on a larger scale. The prosperous owe something to their own countrymen, who fought the nation's wars and built its railways and factories. Sharing wealth among our people binds the classes together, keeps the nation strong, and leaves no divided, resentful populace for rivals to exploit. & 0 & $+1.0$ & $+1.0$ & $+2.0$ & 0 & 0\\[2pt]
            6 & No. Charity is a virtue only when freely given; money seized by the tax collector ennobles no one. Worse, a permanent state dole corrupts the poor themselves, breeding idleness, drink, and dependency, and degrading the self-respect that honest work instills. Relief belongs to churches, families, and neighbors who can guide the needy, not to government clerks. & 0 & $-1.0$ & 0 & 0 & 0 & $+2.0$\\[2pt]
            7 & Yes, and more than that. The wealth of the rich was produced by workers whose labor was paid a fraction of its worth; redistribution merely returns to them what they earned. And the claim is not bounded by flags: the miner in Bolivia and the weaver in Lancashire are robbed by the same owners. & 0 & $+1.0$ & $+2.0$ & $-1.0$ & 0 & 0\\[2pt]
            8 & No. Poverty is painful, but it is nature's spur to industry and prudence. Public relief only encourages the poor to marry early and multiply beyond what the land can feed, so that today's kindness becomes tomorrow's wider misery. The sympathetic impulse must yield to hard economic law; interference does more harm than good. & $-1.0$ & $-1.0$ & 0 & 0 & 0 & 0\\[2pt]
            \bottomrule
            \end{tabular}
            \captionof{table}{Calibration question C3: ``Should the state redistribute wealth to reduce inequality between rich and poor?''}\label{tab:calib-C3}
            \end{center}
            
            \begin{center}\footnotesize
            \begin{tabular}{@{}r p{0.535\linewidth} rrrrrr@{}}\toprule
            \# & Response & Care & Equal. & Prop. & Loyal. & Auth. & Purity\\\midrule
            1 & Yes. My people share my language, my ancestors, my history, and my fate; they are my extended family. A man who claims to love all humanity alike usually loves no one in particular. Charity begins at home, and our first duty is to our own countrymen, before any stranger overseas. & 0 & $-1.0$ & 0 & $+2.0$ & 0 & 0\\[2pt]
            2 & No. A child starving in a distant country suffers exactly as much as a child starving down my street, and the accident of a border does not make her pain matter less. Where we can relieve great suffering cheaply abroad, patriotism is no excuse for looking away and spending lavishly on comforts at home. & $+2.0$ & $+1.0$ & 0 & $-1.5$ & 0 & 0\\[2pt]
            3 & No. Every human being has the same worth and the same rights, whatever passport they hold. Giving my own nation automatic priority is a kind of favoritism no different in principle from favoring my own race or class. Governments may serve their citizens, but morally, no one counts for more because of where they were born. & 0 & $+2.0$ & 0 & $-2.0$ & 0 & 0\\[2pt]
            4 & Yes, though rightly understood it is allegiance to one's sovereign. A subject is bound by oath and by nature to the prince God has set over him and to the laws of the realm. Talk of duties to all mankind is a dreamer's excuse for disobedience; the peace of nations rests on subjects keeping faith with their rulers. & 0 & 0 & 0 & $+1.0$ & $+2.0$ & 0\\[2pt]
            5 & No. My first allegiance belongs to God and to the community of believers scattered across every land. Nations and princes are passing things, and to exalt one's country above the divine law is a form of idolatry that stains the soul. Where my nation commands what God forbids, I must refuse it. & 0 & 0 & 0 & $-1.0$ & $-0.5$ & $+1.5$\\[2pt]
            6 & Yes. My nation schooled me, protected me, built the roads I travel, and sustains the institutions I rely on; its people have paid taxes and fought wars on my behalf. Obligations should track what we have received. Humanity at large, however worthy, has not given me that, so its claim on me is weaker. & 0 & 0 & $+2.0$ & $+1.0$ & 0 & 0\\[2pt]
            7 & Yes. Each people has its own soul, expressed in its tongue, songs, customs, and faith, and that inheritance must be kept whole and undiluted. Cosmopolitan fashions from abroad cheapen and corrupt it, reducing living cultures to a bland, rootless sameness. Guarding our own heritage comes first; humanity is enriched by distinct peoples, not by their dissolution. & 0 & 0 & 0 & $+2.0$ & $+1.0$ & $+2.0$\\[2pt]
            8 & No. Nations are banners that kings and generals wave to send poor men off to kill other poor men for someone else's profit. The worker in one land has more in common with the worker across the border than with his own rulers. I owe the flag nothing; I owe my fellow humans solidarity. & $+0.5$ & $+1.0$ & 0 & $-2.0$ & $-1.0$ & 0\\[2pt]
            \bottomrule
            \end{tabular}
            \captionof{table}{Calibration question C4: ``Should a person's first loyalty be to their own nation and people rather than to humanity at large?''}\label{tab:calib-C4}
            \end{center}
            
            \begin{center}\footnotesize
            \begin{tabular}{@{}r p{0.535\linewidth} rrrrrr@{}}\toprule
            \# & Response & Care & Equal. & Prop. & Loyal. & Auth. & Purity\\\midrule
            1 & Yes. Justice requires that the criminal receive what his deed deserves, measured by the gravity of the crime, not by any benefit to society. Even if a community were about to disband, it should first punish its last murderer. Letting a grave wrong go unanswered because punishment is 'useless' makes a mockery of the victim. & 0 & 0 & $+2.0$ & 0 & 0 & 0\\[2pt]
            2 & Yes. Innocent blood pollutes the land, and the stain remains until the guilty answer for it. A community that shelters a murderer from his due shares in his defilement and invites divine judgment on all. Punishment is not about usefulness; it is the cleansing of a wrong that otherwise lingers among us. & 0 & 0 & $+1.0$ & 0 & 0 & $+2.0$\\[2pt]
            3 & Yes. Even when a given sentence seems pointless, the majesty of the law depends on its being enforced without flinching. When judges soften sentences, the people learn that the crown's commands can be defied, and order begins to unravel. Severe, certain punishment proclaims that the law is sovereign and not to be trifled with. & 0 & 0 & 0 & 0 & $+2.0$ & 0\\[2pt]
            4 & No. Punishment is itself a harm, the deliberate infliction of suffering, and it can only be justified if it prevents greater suffering. Pain that deters no one, protects no one, and reforms no one is simply cruelty added to the world. Talk of what criminals 'deserve' is vengeance dressed up as justice. & $+2.0$ & 0 & $-2.0$ & 0 & 0 & 0\\[2pt]
            5 & No. Vengeance belongs to God, not to us, and Christ commands us to forgive as we have been forgiven. The worst sinner remains a soul capable of repentance, and the aim of any penalty should be his amendment. Heaping suffering on a man for its own sake hardens hearts, his and ours alike. & $+1.0$ & 0 & $-1.0$ & 0 & $+1.0$ & $+0.5$\\[2pt]
            6 & No. Harsh sentences never fall evenly: the poor, the foreign, and the despised fill the prisons and gallows, while the well-connected who commit grave crimes buy lenient terms or escape. A system that punishes severely for its own sake becomes an engine for crushing those at the bottom. Until justice is blind to rank, severity only deepens injustice. & $+1.0$ & $+2.0$ & 0 & 0 & $-0.5$ & 0\\[2pt]
            7 & Yes. When someone murders one of our own, the whole community is wounded, and we owe it to the victim and the grieving family to stand with them. Letting the killer off lightly because punishment is 'pointless' betrays our dead. Whoever turns on the community forfeits its protection and must face its full judgment. & 0 & 0 & $+1.0$ & $+2.0$ & 0 & 0\\[2pt]
            8 & No. There is no stain on society that a prisoner's suffering washes away; that is superstition. Severe punishment mostly serves the state's appetite for displaying its power. Better to focus on repairing harm: bringing offenders face to face with those they hurt, making restitution, and rebuilding lives, rather than caging people to satisfy officials. & $+1.0$ & 0 & $-0.5$ & 0 & $-1.5$ & $-2.0$\\[2pt]
            \bottomrule
            \end{tabular}
            \captionof{table}{Calibration question C5: ``Should those who commit serious crimes be punished severely, even when it does no one any good?''}\label{tab:calib-C5}
            \end{center}
            
            \begin{center}\footnotesize
            \begin{tabular}{@{}r p{0.535\linewidth} rrrrrr@{}}\toprule
            \# & Response & Care & Equal. & Prop. & Loyal. & Auth. & Purity\\\midrule
            1 & No. Marriage is a holy sacrament in which husband and wife become one flesh, and what God has joined no man may put asunder. Separation from bed and board may be allowed for grave cause, but the bond itself endures until death. To remarry while one's spouse lives is adultery and defiles the soul. & 0 & 0 & 0 & 0 & $+1.0$ & $+2.0$\\[2pt]
            2 & Rarely, if ever. The family is the foundation of the social order, with the husband at its head, and loosening the marriage bond loosens every other bond. A wife must bear patiently with her husband's faults, though he may put away an adulterous wife. Easy divorce would unsettle every household and leave society adrift. & $-1.0$ & $-1.5$ & 0 & $+0.5$ & $+2.0$ & 0\\[2pt]
            3 & Yes. No one should be trapped for life with a spouse who beats, terrorizes, or torments them, and children suffer when they grow up amid constant cruelty and fear. Forcing people to remain in such homes causes enormous, needless misery. Divorce is sometimes the only way to protect the vulnerable from ongoing harm. & $+2.0$ & 0 & 0 & 0 & 0 & 0\\[2pt]
            4 & Yes, and on the same terms for wives as for husbands. For centuries a man could cast off an unfaithful wife while she had to endure his infidelities and had no escape. Whatever grounds justify ending a marriage, they must be available to both partners alike, with property and custody divided without favoring either sex. & 0 & $+2.0$ & 0 & 0 & 0 & 0\\[2pt]
            5 & Yes, but only for grave fault such as adultery or willful desertion. The spouse who breaks the covenant has forfeited their claim on it, and the innocent party should be free to remarry. It would be unjust to chain a faithful wife or husband forever to one who has betrayed them, while the offender pays nothing. & 0 & 0 & $+2.0$ & 0 & 0 & 0\\[2pt]
            6 & No. When you marry, you pledge to stand by your spouse in sickness and poverty, through bad years as well as good. Walking away when things become hard is a betrayal of the person who trusted you and of the two families your union joined. Real commitment means staying, especially when it is difficult. & 0 & 0 & 0 & $+2.0$ & 0 & 0\\[2pt]
            7 & Yes, freely. Marriage is a private agreement between two adults, not a sacred mystery for priests or legislators to police. If both, or even one, no longer wish to continue, neither church nor state has any business binding them together. Nobody owes lifelong devotion to a promise made years ago by a different self. & 0 & 0 & 0 & $-1.0$ & $-1.5$ & $-1.0$\\[2pt]
            8 & Only in extreme cases. Whatever the parents feel, children are the ones who pay for divorce, shuttled between homes, torn between mother and father, and often poorer and more anxious for years. Adults who chose to have children should put their own unhappiness second and keep the home together unless there is real danger. & $+2.0$ & 0 & 0 & 0 & 0 & 0\\[2pt]
            \bottomrule
            \end{tabular}
            \captionof{table}{Calibration question C6: ``Should a married couple be permitted to divorce?''}\label{tab:calib-C6}
            \end{center}

    \subsection{Demonstration set}\label{app:demonstration-set}

        \subsubsection{Generation}
            These questions were manually selected from a long list generated by \texttt{Claude Opus 4}.

            For each demonstration question we specified 10 target profiles in $\{-2,\dots,2\}^6$ (Table~\ref{tab:targets}). The profiles were chosen to span traditional and progressive positions and to include cross-cutting ones (for example, protective paternalism: high Care and low Equality). Within each question, some profiles are deliberately identical so that the slate contains approximate clones: Q1 \#1--2 and \#4--6; Q2 \#1--2 and \#3--4; Q3 \#1--3 and \#4--5. The short position labels in Table~\ref{tab:targets} are our design notes and were \emph{not} shown to the generator.

            \begin{table}[h]\centering\scriptsize
                \begin{tabular}{lllrrrrrr}\toprule
                    Q & \# & design note (not shown to generator) & Care & Equal. & Prop. & Loyal. & Auth. & Purity\\\midrule
                    Q1 & 1 & egalitarian rights (A) & $+1$ & $+2$ & 0 & 0 & $-1$ & 0\\
                    Q1 & 2 & egalitarian rights (A, clone) & $+1$ & $+2$ & 0 & 0 & $-1$ & 0\\
                    Q1 & 3 & substantive equality & $+2$ & $+2$ & $-1$ & 0 & $-1$ & 0\\
                    Q1 & 4 & divinely ordained spheres (B) & 0 & $-2$ & $+1$ & $+1$ & $+2$ & $+2$\\
                    Q1 & 5 & divinely ordained spheres (B, clone) & 0 & $-2$ & $+1$ & $+1$ & $+2$ & $+2$\\
                    Q1 & 6 & divinely ordained spheres (B, clone 2) & 0 & $-2$ & $+1$ & $+1$ & $+2$ & $+2$\\
                    Q1 & 7 & rights proportionate to station & 0 & $-2$ & $+2$ & 0 & $+1$ & 0\\
                    Q1 & 8 & civil but not political rights & $+1$ & 0 & $+1$ & $+1$ & $+1$ & 0\\
                    Q1 & 9 & protective paternalism & $+2$ & $-1$ & 0 & $+1$ & $+1$ & $+1$\\
                    Q1 & 10 & communitarian skepticism of rights talk & $+1$ & 0 & 0 & $+2$ & $+1$ & 0\\
                    \midrule
                    Q2 & 1 & partnership of equals (A) & $+1$ & $+2$ & 0 & 0 & $-2$ & $-1$\\
                    Q2 & 2 & partnership of equals (A, clone) & $+1$ & $+2$ & 0 & 0 & $-2$ & $-1$\\
                    Q2 & 3 & holy matrimony, husband as head (B) & 0 & $-2$ & 0 & $+1$ & $+2$ & $+2$\\
                    Q2 & 4 & holy matrimony, husband as head (B, clone) & 0 & $-2$ & 0 & $+1$ & $+2$ & $+2$\\
                    Q2 & 5 & private contract, not for the state & 0 & $+1$ & $+1$ & $-1$ & $-2$ & $-1$\\
                    Q2 & 6 & soft complementarity: headship as duty of provision & $+1$ & $-1$ & $+1$ & $+1$ & $+1$ & $+1$\\
                    Q2 & 7 & lifelong commitment for children's sake, no headship & $+2$ & $+1$ & 0 & $+2$ & 0 & 0\\
                    Q2 & 8 & natural-law procreative ends & $+1$ & $-1$ & 0 & $+1$ & $+1$ & $+2$\\
                    Q2 & 9 & the formula has harmed women and others & $+2$ & $+2$ & 0 & $-1$ & $-1$ & $-2$\\
                    Q2 & 10 & household order as the order of the realm & 0 & $-2$ & $+1$ & $+2$ & $+2$ & 0\\
                    \midrule
                    Q3 & 1 & yes, the magistrate must root it out (B) & $+1$ & 0 & $+1$ & $+1$ & $+2$ & $+2$\\
                    Q3 & 2 & yes, the magistrate must root it out (B, clone) & $+1$ & 0 & $+1$ & $+1$ & $+2$ & $+2$\\
                    Q3 & 3 & yes, the magistrate must root it out (B, clone 2) & $+1$ & 0 & $+1$ & $+1$ & $+2$ & $+2$\\
                    Q3 & 4 & no, superstition that kills innocents (A) & $+2$ & $+1$ & 0 & 0 & $-1$ & $-1$\\
                    Q3 & 5 & no, superstition that kills innocents (A, clone) & $+2$ & $+1$ & 0 & 0 & $-1$ & $-1$\\
                    Q3 & 6 & prosecute only real underlying crimes & $+2$ & 0 & $+2$ & 0 & $+1$ & $-1$\\
                    Q3 & 7 & a matter for church penance, not the civil sword & $+1$ & 0 & 0 & $+1$ & $+1$ & $+2$\\
                    Q3 & 8 & medical and psychological care instead & $+2$ & 0 & $-1$ & 0 & 0 & $-2$\\
                    Q3 & 9 & due process and liberty of conscience forbid it & $+1$ & $+2$ & 0 & 0 & $-2$ & $-2$\\
                    Q3 & 10 & punish to protect the community, mercy for the penitent & $+1$ & 0 & $+1$ & $+2$ & $+2$ & $+1$\\
                \bottomrule\end{tabular}
                \caption{Design target profiles for the demonstration responses. Identical rows within a question are intended near-clones.}\label{tab:targets}
            \end{table}

            The responses to each demonstration question were generated in a separate call. The prompt for Q1 is reproduced below; the prompts for Q2 and Q3 differ only in the question and the list of targets (Table~\ref{tab:targets}).

            \begin{lstlisting}[numbers=none,basicstyle=\ttfamily\scriptsize,breaklines=true,columns=fullflexible,frame=single,aboveskip=6pt,belowskip=6pt]
You are helping construct a clearly-labelled SYNTHETIC dataset for a research paper on aggregating diverse and historically changing moral views with voting rules. Your job is to write candidate answers to a question. The answers will be presented as synthetic in the paper.

We describe moral outlooks using the six foundations of the Moral Foundations Questionnaire-2 (Atari et al., 2023):
- Care: Intuitions about avoiding emotional and physical damage to another individual.
- Equality: Intuitions about equal treatment and equal outcome for individuals.
- Proportionality: Intuitions about individuals getting rewarded in proportion to their merit or contribution.
- Loyalty: Intuitions about cooperating with ingroups and competing with outgroups.
- Authority: Intuitions about deference toward legitimate authorities and the defense of traditions, all of which are seen as providing stability and fending off chaos.
- Purity: Intuitions about avoiding bodily and spiritual contamination and degradation.

For each target below, write ONE candidate answer to the question whose stance and reasoning match the target profile. A profile gives, for each foundation, how strongly the answer should uphold (+2 strongly, +1 somewhat) or go against / subordinate (-1 somewhat, -2 strongly) that concern, or leave it unengaged (0).

Requirements:
- 35 to 60 words each, written as a direct answer to the question, in the voice of someone sincerely holding that view. Views should be ones that people have plausibly held at some point between the 16th and 21st centuries.
- Do NOT name the foundations or use the words care, equality, proportionality, loyalty, authority, or purity (or their obvious derivatives such as "equal", "loyal", "pure", "authoritative"); convey the stance through content and reasoning instead. (Other people will later rate your answers on these dimensions without seeing the targets.)
- Where two or more targets have identical profiles, write genuinely different wordings of essentially the same position (a paraphrase with different phrasing and examples, not a copy).
- No slurs, no calls to violence against real groups; historical views may be illiberal but should be expressed as their holders would have argued them.

Question: "Should women have the same legal and political rights as men?"

Targets:
1. Care +1, Equality +2, Proportionality +0, Loyalty +0, Authority -1, Purity +0
2. Care +1, Equality +2, Proportionality +0, Loyalty +0, Authority -1, Purity +0
3. Care +2, Equality +2, Proportionality -1, Loyalty +0, Authority -1, Purity +0
4. Care +0, Equality -2, Proportionality +1, Loyalty +1, Authority +2, Purity +2
5. Care +0, Equality -2, Proportionality +1, Loyalty +1, Authority +2, Purity +2
6. Care +0, Equality -2, Proportionality +1, Loyalty +1, Authority +2, Purity +2
7. Care +0, Equality -2, Proportionality +2, Loyalty +0, Authority +1, Purity +0
8. Care +1, Equality +0, Proportionality +1, Loyalty +1, Authority +1, Purity +0
9. Care +2, Equality -1, Proportionality +0, Loyalty +1, Authority +1, Purity +1
10. Care +1, Equality +0, Proportionality +0, Loyalty +2, Authority +1, Purity +0

Return ONLY a JSON list, with no other text, of the form
[{"target_id": 1, "text": "..."}, ...]
with one entry per target, in order.

Answer directly, without using any tools.
\end{lstlisting}

        \subsubsection{Questions}
            Tables~\ref{tab:demo-Q1}--\ref{tab:demo-Q3} list the three demonstration questions and their responses, with each response's feature vector $\phi(i)$. Response numbers match Table~\ref{tab:targets}, Table~\ref{tab:synth_votes}, and Figures~\ref{fig:values} and~\ref{fig:winners}.

            \begin{center}\footnotesize
            \begin{tabular}{@{}r p{0.535\linewidth} rrrrrr@{}}\toprule
            \# & Response & Care & Equal. & Prop. & Loyal. & Auth. & Purity\\\midrule
            1 & Yes. A woman is as much a person as any man, and the law should treat her so: the same vote, the same right to own property, sign contracts, serve on juries, and hold office. Customs that barred her were written by men for men; age alone doesn't make them just, and they have left countless women trapped and suffering. & $+1.0$ & $+2.0$ & 0 & 0 & $-1.0$ & 0\\[2pt]
            2 & Absolutely. Rights belong to persons, not to one sex. Women should vote, inherit, testify, stand for parliament, and keep their own wages on exactly the terms their brothers do. That clergy and old statute books once said otherwise proves only that they erred, and the error has cost women dearly in poverty and helplessness. & $+1.0$ & $+2.0$ & 0 & 0 & $-1.0$ & 0\\[2pt]
            3 & Yes, the very same rights men hold, without conditions. Denying women rights has meant beaten wives with no escape, mothers stripped of their children, girls married off against their will. Rights shouldn't have to be earned by proving usefulness or achievement; every woman is owed them simply as a human being. Traditions that sanctioned such suffering deserve no deference. & $+2.0$ & $+2.0$ & $-1.5$ & 0 & $-1.5$ & 0\\[2pt]
            4 & No. Scripture and the settled order of every Christian nation make the husband head of the household, and the realm is built of such households. Dragging women into the grime of electioneering would sully the modesty that keeps the home holy. Those who bear arms and toil for the country rightly govern it. & 0 & $-2.0$ & $+1.0$ & 0 & $+2.0$ & $+1.0$\\[2pt]
            5 & Certainly not. God made man and woman for different stations, and our fathers' wisdom has kept each in its place. A woman's honour lies in a chaste, sheltered life, not the corrupting clamour of the hustings. Men who defend the nation and provide for its families have earned the franchise; scattering it freely would unravel our people's order. & 0 & $-2.0$ & $+1.0$ & 0 & $+2.0$ & $+1.0$\\[2pt]
            6 & No. Heaven set the man over the woman as Christ over the Church, and a kingdom overturning that order invites chaos. Separate votes and estates for wives would pit them against husbands and split the family our community rests upon. Keeping women clear of politics' coarseness preserves their virtue; governing belongs to those who shoulder its burdens. & 0 & $-2.0$ & $+1.0$ & $+1.0$ & $+2.0$ & $+1.0$\\[2pt]
            7 & No. Political power should follow contribution. Men pay the bulk of taxes, serve in the militia, and bear the costs of war; those who carry the burdens of the state have earned the right to steer it. Women, who contribute differently, should not claim an identical share. Our constitution has long tied the franchise to such duties, and wisely. & 0 & $-2.0$ & $+2.0$ & 0 & $+1.0$ & 0\\[2pt]
            8 & Women who kept our munitions factories and hospitals running through the war have earned a voice in the nation they served. Let Parliament extend the franchise to them in orderly, lawful steps rather than yielding to street agitation, and let their votes help protect widows, children, and the working poor. & $+1.0$ & $+1.0$ & $+2.0$ & $+1.0$ & $+1.0$ & 0\\[2pt]
            9 & Not identical rights, no. Women and children need shielding from the harshness of factory, courtroom, and campaign, and laws that treat them just like men would strip away the protections that keep them from exploitation. Let husbands and fathers speak for their households as custom intends, and let women's gentler virtues stay unspoiled, safeguarding families and the community we share. & $+1.5$ & $-2.0$ & 0 & $+1.0$ & $+1.0$ & $+1.0$\\[2pt]
            10 & Our people cannot throw off foreign rule while half our strength sits idle. The movement needs its daughters as much as its sons, so let our national leaders write women's votes into the new constitution. Every household will then stand united behind the cause, and our mothers can guard our children's welfare against the occupier's laws. & $+1.0$ & $+1.0$ & 0 & $+2.0$ & 0 & 0\\[2pt]
            \bottomrule
            \end{tabular}
            \captionof{table}{Demonstration question Q1: ``Should women have the same legal and political rights as men?''}\label{tab:demo-Q1}
            \end{center}
            
            \begin{center}\footnotesize
            \begin{tabular}{@{}r p{0.535\linewidth} rrrrrr@{}}\toprule
            \# & Response & Care & Equal. & Prop. & Loyal. & Auth. & Purity\\\midrule
            1 & No. Marriage should be a partnership between two consenting adults, whatever their sexes, who share decisions, money, and chores as peers. Giving husbands command merely because they are men has trapped countless wives in misery. Neither church nor state should decree who rules at home, and old taboos about which couples are 'natural' deserve no place in law. & $+1.0$ & $+2.0$ & 0 & 0 & $-1.0$ & $-1.0$\\[2pt]
            2 & Certainly not. The law should never crown husbands as rulers of the home; coverture stripped wives of property and voice for centuries. Whoever the partners are, man and woman or two of the same sex, each deserves an identical say, and anyone trapped in a cruel match should be free to leave despite clerics' talk of sacred permanence. & $+1.0$ & $+2.0$ & 0 & 0 & $-1.0$ & $-1.0$\\[2pt]
            3 & Yes. God ordained marriage in Eden as one man and one woman, made one flesh until death, and Scripture commands wives to submit to their husbands as the Church submits to Christ. To level husband and wife into interchangeable partners is to overturn the order He established and to defile a holy sacrament our people have kept for generations. & 0 & $-1.5$ & 0 & 0 & $+2.0$ & $+1.5$\\[2pt]
            4 & It must be so. Every household is a little commonwealth, and as the king governs the realm, the husband governs wife, children, and servants. Women are not fitted by nature for rule. Marriage is a holy estate, and fornication, divorce, and unnatural couplings pollute both the body and the congregation; a godly people must keep that estate unspotted. & 0 & $-2.0$ & 0 & 0 & $+2.0$ & $+2.0$\\[2pt]
            5 & No. Marriage is a private contract between two individuals, not a sacrament and not a duty owed to one's clan, parish, or nation. Nobody should be handed command by accident of sex; if one spouse earns the larger say by bringing more skill or labour to the household, let that be negotiated, not decreed by custom or clergy. & 0 & $+1.0$ & $+1.5$ & $-1.0$ & $-1.0$ & $-1.0$\\[2pt]
            6 & Yes. A husband who toils to provide for and shield his wife and children has earned the final word in family matters, as she earns honour for running the home. Such lasting, God-blessed unions keep children secure, bind families to their neighbours, and follow the wisdom our grandparents handed down, though a good head listens before he decides. & $+1.0$ & $-1.0$ & $+2.0$ & $+1.0$ & $+1.0$ & 0\\[2pt]
            7 & What matters is not who heads the house but that spouses stand by each other through sickness, poverty, and grief, and never abandon their children or kin. A marriage should be a shelter where nobody is hurt, where both have a voice, and where the family closes ranks against whatever hardship comes. & $+1.0$ & $+1.0$ & 0 & $+2.0$ & $-1.0$ & 0\\[2pt]
            8 & Yes. Marriage is a sacred covenant that hallows the union of man and woman and guards against lust, adultery, and the degradation of the body. The husband's headship, exercised gently, gives the home settled order and shelters wife and children from harm. Man and wife have different callings, and homes built this way keep our congregation strong. & $+1.0$ & $-1.0$ & 0 & $+1.0$ & $+1.0$ & $+2.0$\\[2pt]
            9 & No. Tying marriage to one man and one woman forever, with him in charge, has chained people to abusive spouses and shamed gay couples and single mothers. Sex and love are not sins needing a license. Adults should be free to marry, divorce, or stay single with identical rights, and no one should endure misery to spare the family's name. & $+1.5$ & $+1.5$ & 0 & $-1.0$ & $-0.5$ & $-1.5$\\[2pt]
            10 & Yes. A household, like an army, needs one commander, and the husband earns that post by shouldering its debts, defence, and good name. Wife and children owe him obedience, as he owes his clan and nation service. Granting every member the same say would dissolve the family into squabbling strangers. & 0 & $-2.0$ & $+1.0$ & $+1.0$ & $+2.0$ & 0\\[2pt]
            \bottomrule
            \end{tabular}
            \captionof{table}{Demonstration question Q2: ``Should marriage be defined as a lifelong union between a man and a woman in which the husband is the head of the household?''}\label{tab:demo-Q2}
            \end{center}
            
            \begin{center}\footnotesize
            \begin{tabular}{@{}r p{0.535\linewidth} rrrrrr@{}}\toprule
            \# & Response & Care & Equal. & Prop. & Loyal. & Auth. & Purity\\\midrule
            1 & Yes. The witch renounces her baptism and gives herself to Satan, defiling her soul and bringing filth into a Christian parish that must stand together against the Enemy. God ordained magistrates and placed the sword in their hands for such crimes. Neighbours whose children she harmed deserve justice, and she deserves the penalty her wickedness has earned. & $+1.0$ & 0 & $+1.0$ & $+1.0$ & $+1.0$ & $+2.0$\\[2pt]
            2 & Certainly. Witchcraft is the foulest of crimes: a covenant with the Devil that stains the soul and corrupts every household it touches. The prince and his judges are God's lieutenants; leaving such treachery against Heaven and our commonwealth unpunished invites wrath upon us all. She must answer for every blighted crop and sickened infant. & $+0.5$ & 0 & $+1.0$ & $+1.0$ & $+1.5$ & $+2.0$\\[2pt]
            3 & It is the plain duty of the civil magistrate. Whoever consorts with evil spirits makes their body the Devil's vessel and stains our covenanted community, which must stand united against him. Rulers bear the sword by God's appointment and must not flinch. The afflicted girls deserve relief, and so grievous a sin its due wages. & $+1.0$ & 0 & $+1.0$ & $+1.0$ & $+1.5$ & $+2.0$\\[2pt]
            4 & No. Those accused are nearly always poor, lonely old women, blamed for misfortunes they never caused, and racked until they confess to anything. Their suffering is real; their crimes are imagined. They deserve the same fair hearing as any wealthy man. Learned judges and preachers have been wrong before, and tales of devilish taint are fancies, not evidence. & $+2.0$ & $+1.5$ & 0 & 0 & $-1.0$ & $-1.0$\\[2pt]
            5 & The state should stop these trials. I have watched a friendless, penniless widow kept from sleep for days until she named her neighbours, who then suffered in turn. A beggar's life counts no less than a lord's. Let us not blindly trust the witch-finders and clergy, nor their superstitious talk of souls befouled by Satan. & $+2.0$ & $+1.5$ & 0 & 0 & $-1.0$ & $-1.0$\\[2pt]
            6 & Only for real deeds. Spells harm nobody, and talk of spiritual pollution is no ground for torturing the innocent. But one who poisons a neighbour's well, or cheats a grieving mother to 'lift a curse', should answer before the lawful courts, not a village mob, and receive a sentence matched to the actual injury, no more and no less. & $+1.0$ & 0 & $+2.0$ & 0 & $+1.0$ & $-1.0$\\[2pt]
            7 & Yes. Witchcraft is uncleanness of the soul: unholy rites, bodily defilement, commerce with foul spirits, and such corruption spreads through a village like rot through a barrel. We are one people under God and must keep our flock clean for the sake of our children. The magistrate rightly acts to root it out and restore the parish to holiness. & 0 & 0 & 0 & $+1.0$ & $+1.0$ & $+2.0$\\[2pt]
            8 & No. Whatever a woman mutters over herbs, it wounds no one, and the idea that her body or soul is defiled and must be scourged clean is a barbarous superstition. Even if she meant ill, answering with the rack and the stake only multiplies misery. Mercy and gentle help do more good than any tally of what she deserves. & $+2.0$ & 0 & $-1.0$ & 0 & 0 & $-2.0$\\[2pt]
            9 & Absolutely not. Witch trials are how churchmen and magistrates keep women, the poor, and dissenters in line; the accused should have exactly the rights of any lord or bishop. Fretting over sinful female bodies and tainted souls is bigotry dressed as holiness. No one owes obedience to institutions that torture healers and midwives to guard their power. & $+1.0$ & $+2.0$ & 0 & 0 & $-2.0$ & $-2.0$\\[2pt]
            10 & Yes. Witches are the Devil's sworn servants, enemies of the King and Christendom, conspiring to raise storms against His Majesty's ship and ruin the realm. Such foul treason betrays crown, kin and countrymen. The anointed King and his judges hold their charge from God; they must defend his subjects and give these traitors the punishment their treachery merits. & $+0.5$ & 0 & $+1.0$ & $+2.0$ & $+2.0$ & 0\\[2pt]
            \bottomrule
            \end{tabular}
            \captionof{table}{Demonstration question Q3: ``Should those accused of witchcraft be tried and punished by the state?''}\label{tab:demo-Q3}
            \end{center}

\section{Proof-of-concept experiment methodology}\label{app:exp-methods}

    This appendix describes how the results in Section~\ref{sec:exp} were produced. 
    Every analysis uses a fixed random seed.

    \subsection{Reward basis construction}\label{app:basis}
        We use as reward bases the six foundations of the revised Moral Foundations Questionnaire (MFQ-2; \citealp{atari2023morality}). The definitions below are taken verbatim from \citet{atari2023morality}, and are the ones given to both the generator and the raters:
        \begin{itemize}
            \item \textbf{Care}: ``Intuitions about avoiding emotional and physical damage to another individual.''
            \item \textbf{Equality}: ``Intuitions about equal treatment and equal outcome for individuals.''
            \item \textbf{Proportionality}: ``Intuitions about individuals getting rewarded in proportion to their merit or contribution.''
            \item \textbf{Loyalty}: ``Intuitions about cooperating with ingroups and competing with outgroups.''
            \item \textbf{Authority}: ``Intuitions about deference toward legitimate authorities and the defense of traditions, all of which are seen as providing stability and fending off chaos.''
            \item \textbf{Purity}: ``Intuitions about avoiding bodily and spiritual contamination and degradation.''
        \end{itemize}
        The basis set $V$ consists of the six one-hot vectors in $\mathbb{R}^6$, one per foundation, so $V$ is the identity matrix. A response $i$ is represented by a feature vector $\phi(i)$ (described below). Annotator $n$'s reward function is $R_n(i)=w_n^\top V\phi(i)=w_n^\top\phi(i)$, with unconstrained weights $w_n\in\mathbb{R}^6$. Stage~1 therefore consists of fixing these interpretable bases rather than learning them from $\mathcal{D}_0$.

    \subsection{Response feature scoring}\label{app:feature-scoring}
        All 78 responses were shuffled into a single list with a fixed seed, and relabelled R01--R78. Two raters then rated every response. Each rater was a separate, fresh instance of \texttt{Claude Opus 5}. Each saw only the prompt below. They did not see the target profiles, the clone structure, the generation prompts, or each other's ratings. The list of 78 items is omitted here; each item appears in the prompt as \texttt{Question: "...", Answer: "..."}. A response's feature vector $\phi(i)$ is the mean of the two raters' integer ratings, so its entries are multiples of $0.5$.

        \begin{lstlisting}[numbers=none,basicstyle=\ttfamily\scriptsize,breaklines=true,columns=fullflexible,frame=single,aboveskip=6pt,belowskip=6pt]
You are an expert annotator for a research study in moral psychology. You will rate short answers to moral questions on the six foundations of the Moral Foundations Questionnaire-2 (Atari et al., 2023):
- Care: Intuitions about avoiding emotional and physical damage to another individual.
- Equality: Intuitions about equal treatment and equal outcome for individuals.
- Proportionality: Intuitions about individuals getting rewarded in proportion to their merit or contribution.
- Loyalty: Intuitions about cooperating with ingroups and competing with outgroups.
- Authority: Intuitions about deference toward legitimate authorities and the defense of traditions, all of which are seen as providing stability and fending off chaos.
- Purity: Intuitions about avoiding bodily and spiritual contamination and degradation.

For each answer, rate how strongly the answer's STANCE AND REASONING, with respect to the question it answers, upholds or goes against each foundation, on this signed scale:
  +2 = strongly upholds or relies on this concern
  +1 = somewhat upholds it
   0 = does not engage this concern, or is neutral about it
  -1 = somewhat goes against, dismisses, or subordinates it
  -2 = strongly goes against or rejects it
Rate each answer independently of the others, and rate each foundation independently of the other foundations. Judge what the answer says and implies, not whether you agree with it. Answers are presented in random order.

Answers:
R01. Question: "Should a person's first loyalty be to their own nation and people rather than to humanity at large?"
    Answer: "Yes. My people share my language, my ancestors, my history, and my fate; they are my extended family. A man who claims to love all humanity alike usually loves no one in particular. Charity begins at home, and our first duty is to our own countrymen, before any stranger overseas."

[... R02 to R78 in the same format ...]

Return ONLY a JSON object, with no other text, mapping each answer id to a list of six integers in the order [Care, Equality, Proportionality, Loyalty, Authority, Purity], e.g. {"R01": [1, 0, -1, 0, 2, 0], ...}. Include every answer id exactly once.

Answer directly, without using any tools.
\end{lstlisting}

        \paragraph{Agreement.}
            \begin{itemize}
                \item \emph{Between the two raters}, over all $78\times6$ ratings: exact agreement 0.934, agreement within one point 1.000, Pearson $r=0.968$, and linearly weighted Cohen's $\kappa=0.937$. Per foundation, $r$ ranges from 0.94 (Care) to 0.99 (Loyalty).
                \item \emph{Between the design targets and the mean ratings} of the 30 demonstration responses: Pearson $r=0.944$, and the signs agree for 93.7\% of nonzero targets.
            \end{itemize}

    \subsection{Historical annotator simulation}\label{app:annotator-simulation}

        \paragraph{Cohorts.}
            Each period's population is a two-component Gaussian mixture over weight vectors $w\in\mathbb{R}^6$. Table~\ref{tab:cohorts} gives the component means and mixing shares. Each component has isotropic standard deviation $0.3$. We specified these profiles by hand. They loosely encode the broad qualitative claim that the binding foundations (Loyalty, Authority, Purity) and Proportionality received more weight, and Equality less, in earlier periods. Each cohort is a mixture rather than a caricature, so that it has internal diversity. The profiles are not estimated from historical data, so the winner chosen by a homogeneous cohort largely follows by construction.

        \paragraph{Preference data.}
            Each simulated annotator with true weights $w$ answers $m=20$ preference comparisons. Each comparison is a pair of responses drawn uniformly at random, with replacement across comparisons, from the $6\times\binom{8}{2}=168$ within-question pairs of calibration responses. The annotator's choice is drawn from the Bradley--Terry model (Equation~\ref{eq:bolt}) with $\beta=1$.

        \paragraph{Reward model personalization.}
            We estimate each annotator's weights $\hat w$ from their $m$ comparisons by $\ell_2$-regularized maximum likelihood, with $\lambda=0.5$ (Section~\ref{sec:exp}). The objective is strictly convex, and we minimize it with Newton's method, from $w=0$, until the largest step is below $10^{-8}$ (at most 50 iterations). We use the regularizer because, with only 20 comparisons, an annotator's choices can be linearly separable, in which case the unregularized maximum-likelihood estimate diverges. All jurors in the voting experiments use their \emph{estimated} weights $\hat w$, so the votes include estimation error.

    \subsection{Voting}\label{app:voting}

        \paragraph{Juries}
            Each jury has 30 members: 30 from one period for the single-cohort juries, or 10 from each period for the mixed jury. Every draw of a jury samples new annotators and re-fits their weights. Each juror ranks the 10 responses to a demonstration question by $\hat R_n(i)=\hat w_n^\top\phi(i)$. 
            
        \paragraph{Vote aggregation}
            We aggregate the resulting profile with \texttt{pref\_voting} v1.18.2 \citep{Holliday2025-ew}, using plurality, Borda, instant runoff with parallel-universe tie-breaking (IRV-PUT), and Copeland as our voting method. IRV-PUT and Copeland can return several winners, so we break ties by lowest response number.


\end{document}